\documentclass{article}

\PassOptionsToPackage{numbers, compress}{natbib}
 \usepackage[main, final]{neurips_2026}

\usepackage[utf8]{inputenc} % allow utf-8 input
\usepackage[T1]{fontenc}    % use 8-bit T1 fonts
\usepackage{hyperref}       % hyperlinks
\usepackage{url}            % simple URL typesetting
\usepackage{booktabs}       % professional-quality tables
\usepackage{amsfonts}       % blackboard math symbols
\usepackage{nicefrac}       % compact symbols for 1/2, etc.
\usepackage{microtype}      % microtypography
\usepackage{xcolor}         % colors
\usepackage{soul}           % Provides the \hl command
\usepackage{graphicx}
\usepackage{amsmath}
\usepackage{ragged2e, array}
\usepackage{multirow}
\usepackage{subcaption}
\usepackage{wrapfig}
\usepackage{booktabs}
\usepackage{array}
\usepackage{makecell}

\title{Wavelet Flow Matching for Multi-Scale Physics Emulation}

\author{%
  Gabriele Accarino\textsuperscript{1}\thanks{Corresponding author: \texttt{ga2673@columbia.edu}} \quad
  Juan Nathaniel\textsuperscript{1} \quad          % verify affiliation
  Carla Roesch\textsuperscript{2} \quad
  Pierre Gentine\textsuperscript{1,3} \\
  \textbf{Sara Shamekh\textsuperscript{4}} \quad
  \textbf{Duncan Watson-Parris\textsuperscript{5}} \quad
  \textbf{Viviana Acquaviva\textsuperscript{6,7}} \\
  [0.8em]
  \small\textsuperscript{1}Department of Earth and Environmental Engineering, Columbia University, New York, NY, USA \\
  \small\textsuperscript{2}University of Edinburgh, Edinburgh, Scotland, UK \\
  \small\textsuperscript{3}École Normale Supérieure – PSL, Paris, France \\
  \small\textsuperscript{4}Courant Institute of Mathematical Sciences, New York University, New York, NY, USA \\
  \small\textsuperscript{5}Scripps Institution of Oceanography \& Hal\i c\i o\u{g}lu Data Science Institute, UC San Diego, La Jolla, CA, USA \\
  \small\textsuperscript{6}CUNY New York City College of Technology, Brooklyn, NY, USA \\
  \small\textsuperscript{7}Lamont-Doherty Earth Observatory, Columbia University, Palisades, NY, USA \\
}

\begin{document}

\maketitle
\begin{abstract}
%Accurate emulation of multi-scale physical systems governed by PDEs demands models that remain stable over long autoregressive rollouts while preserving fine-scale structures. Deterministic emulators produce overly-smoothed predictions, while generative approaches better capture details but are costly. Latent-space generative models have emerged as a compromise but with the additional cost of separately pre-trained autoencoders. We propose \textbf{Wavelet Flow Matching} (WFM), a novel generative emulator that overcomes current trade-offs between cost and skill by performing optimal-transport directly in the multi-scale wavelet space. Rather than learning a latent compression, WFM leverages the hierarchical structure of a U-Net to jointly predict transport velocities of a prescribed wavelet representation. On three challenging systems of chaotic fluid dynamics, WFM achieves superior long-horizon stability, accuracy and spectral coherence compared to state-of-the-art models. Our results clearly position the wavelet space as an effective training-free representation for generative emulation of complex physical dynamics.

Accurate emulation of multi-scale physical systems governed by PDEs demands models that remain stable over long autoregressive rollouts while preserving fine-scale structures. Deterministic emulators produce overly smoothed predictions, while generative approaches better capture details but are costly. Latent-space generative models have emerged as a compromise but with the additional cost of separately pre-trained autoencoders. We propose \textbf{Wavelet Flow Matching} (WFM), a generative emulator that overcomes current trade-offs between cost and skill by performing optimal-transport flow matching directly in the multi-scale wavelet space. Rather than learning a latent compression, WFM leverages the hierarchical structure of a U-Net to jointly predict the transport velocities of a prescribed, lossless wavelet representation. On three challenging systems of chaotic fluid dynamics, WFM achieves superior spectral coherence and competitive or better long-horizon accuracy compared to state-of-the-art models. Our results position the wavelet space as an effective training-free representation for generative emulation of complex physical dynamics.
\end{abstract}

\section{Introduction}\label{sec:introduction}
Simulating complex physical systems governed by partial differential equations (PDEs) lies at the heart of scientific computing, with applications spanning turbulence modeling ~\citep{pope2000}, weather and climate prediction ~\citep{bi2023,nathaniel2024chaosbench,lam2023}, and astrophysics~\citep{ohana2024}. Classical numerical solvers are accurate but computationally prohibitive when thousands of trajectories need to be generated ~\citep{palmer2019, kochkov2021}. This has motivated a growing body of work on data-driven emulators trained to predict future system states from past ones at a fraction of the cost of high-fidelity simulations ~\citep{li2021, mccabe2024}. 

The canonical formulation treats emulation as an autoregressive prediction task: given the current state $x^t$ (and optionally a context window of $L$ past states $x^{t-L:t-1}$), a neural network $f_\theta$ is trained to predict the next state $x^{t+1}$. Deterministic emulators such as variants of Neural Operators (NOs)~\citep{li2021, kossaifi2023tfno, raonic2023convolutional}, U-Nets~\citep{ronneberger2015}, and transformers~\cite{li2023}, are trained by regression loss and are extraordinarily fast at inference. However, squared-error training encourages over-smoothed predictions, and, crucially, small errors compound over autoregressive rollouts. At each rollout step, the model receives input that can be out of the data manifold on which it has been trained, progressively drifting from the true dynamics~\citep{mccabe2023,lippe2023}. This effect is especially severe for chaotic systems, where trajectories diverge exponentially~\citep{kohl2024}.

Generative models address this failure mode by learning the full conditional distribution $p(x^{t+1}\!\mid\!x^t)$ rather than its mean~\citep{accarino2026learning}. Sampling from it introduces structured stochasticity and mitigates rollout drift~\citep{rombach2022, lippe2023, kohl2024, shysheya2024, huang2024diffusionpde, rozet2025, nathaniel2026generative}. Among generative frameworks, Flow Matching (FM)~\citep{lipman2023, lipman2024, liu2023, tong2024} is particularly attractive, as it learns a continuous flow by regressing a conditional vector field. This approach achieves high sample quality on turbulent PDE systems while maintaining high inference efficiency~\citep{li2025}, making it a competitive approach for generative modeling~\citep{gupta2025}.

\begin{figure}[t]
  \centering
  %\fbox{\rule[-.5cm]{0cm}{4cm} \rule[-.5cm]{4cm}{0cm}}
  \includegraphics[width=1.0\textwidth]{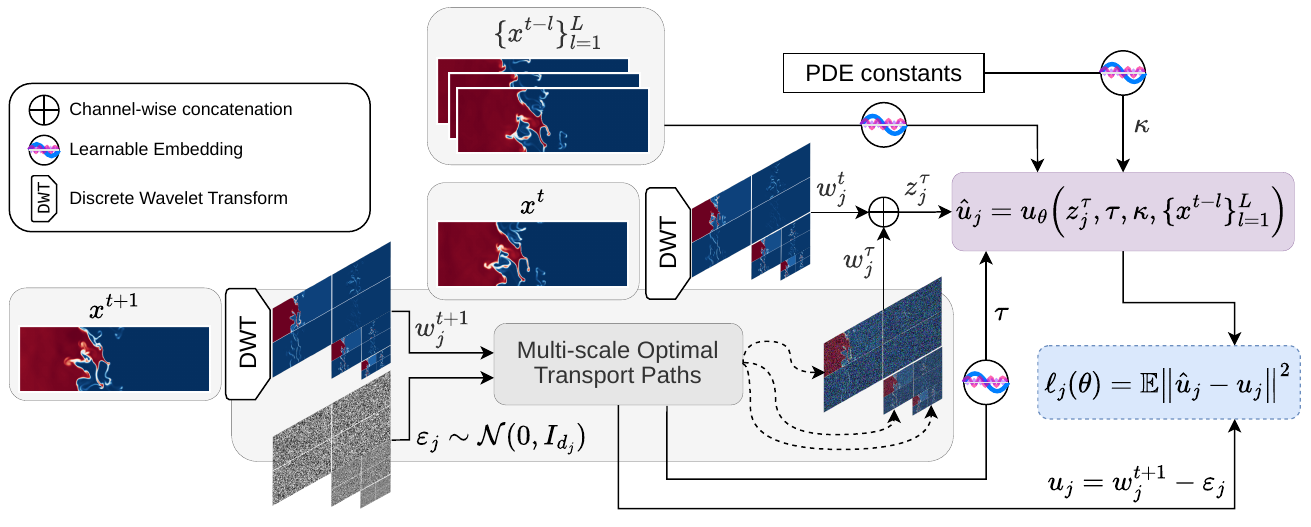}
  \caption{Illustration of the Wavelet Flow Matching architecture. During training, Gaussian noise is sampled to match the shape of wavelet coefficients at each scale. OT-CFM interpolation is applied between the next-state wavelet coefficients \(w_j^{t+1}\) and noise at every scale with a shared interpolation time \(\tau\), yielding \(J\) per-scale flows that are coupled through the shared U-Net. The velocity-predicting U-Net \(u_{\theta}\) (purple box) is conditioned on the interpolation step \(\tau\), the channel-wise concatenation $z_j^{\tau}$ of \(w_j^{\tau}\) and the current-state coefficients \(w_j^{t}\), PDE constants \(\kappa\), and embedding of past $L$ frames. The illustration of $u_{\theta}$ is shown in Figure~\ref{fig:unet_diagram}.}
  \label{fig:graphical_abstract}
\end{figure}

Recent works on representation-based FM for physics emulation have shown great success. Latent-space formulations apply FM in the space of learned autoencoders~\cite{dao2023flow}, reducing computational cost but introducing additional training overhead and are less interpretable. This has motivated the use of physics-based representations, most prominently Fourier approaches~\citep{kossaifi2023tfno, li2024}, where FM is performed in a fixed spectral basis that can generalize across resolutions. However, such representations are inherently global and do not explicitly capture localized multi-scale structure, which is central to many physical systems. As such, we ask the following question:

\begin{center}
\emph{Are there representations in which the generative process can be carried out naturally across multiple scales, while preserving accuracy, spectral coherence, and physical plausibility of the emulated dynamics?} 
\end{center}

%We answer by proposing to perform the generative process, particularly FM, in the space of wavelet coefficients. 
%as the task of generating the multi-scale wavelet representation of the next-state.

As an answer to this question, in this work we present \textbf{Wavelet Flow Matching} (WFM), a framework that performs physical emulation in the wavelet space. WFM leverages Optimal-Transport Conditional Flow Matching (OT-CFM)~\citep{tong2024} that performs autoregressive rollouts on the multi-scale wavelet representation of physical fields. Discrete Wavelet Transform (DWT) representations are grounded in classical harmonic analysis, require no training, and are simultaneously localized in space (or time) and scale~\citep{mallat2009}, i.e., capturing both where frequencies occur and at what resolution. Each scale of the DWT reduces the spatial resolution, so that coarser coefficient maps live on progressively smaller grids. Training and sampling are carried out entirely in wavelet space, while an Inverse DWT (IDWT) is applied to the generated coefficients only at the end of the sampling procedure to reconstruct the next physical state (Figure~\ref{fig:graphical_abstract}). The framework is highly configurable, for instance users can choose the desired number of scales, as well as different types of prescribed wavelet representations.
\clearpage
Our \textbf{main contributions} are (Figure~\ref{fig:graphical_abstract}):
\begin{itemize}
    %\item We re-frame physical emulation as conditional generation of next-state multi-scale wavelet coefficients;
    \item We introduce WFM, a framework performing autoregressive emulation in the wavelet space, leveraging the hierarchical structure of U-Nets to predict multi-scale flow velocities.
    
    \item We benchmark WFM on three challenging PDE datasets, showing that WFM outperforms other baselines in terms of spectral coherence RMSE and overall long-range skill. The advantage of multi-scale wavelet representation is especially evident in the most chaotic regimes.  
    
    %\item WFM shows favorable speedup with no extra cost of a prior learned compression.
    \item WFM offers a favourable accuracy--cost trade-off: it needs no pre-trained compression and samples 1.1--1.5$\times$ faster than pixel-space flow matching at comparable or better accuracy.
\end{itemize}

To facilitate further research in this direction, we publicly release our code at \url{https://github.com/gabrieleaccarino/wavelet_flow_matching}, along with the pre-trained model weights at \url{https://huggingface.co/gabrieleaccarino/wavelet-flow-matching}.

\section{Background}\label{sec:background}
\subsection{Optimal-Transport Conditional Flow Matching}\label{sec:ot-cfm}

Let $p_0 = \mathcal{N}(0, I_d)$ denote a simple source distribution and let $p_1$ be the target distribution on $\mathbb{R}^d$, accessed through $n$ samples $x_1^{(1)}, \dots, x_1^{(n)} \sim p_1$. The goal of FM is to learn a time-dependent velocity field $u : \mathbb{R}^d \times [0,1] \to \mathbb{R}^d$ that transports samples from $p_0$ to $p_1$.

This transport is defined through the ordinary differential equation (ODE):
\begin{equation}
    \begin{cases}
        x(0) = x_0 \sim p_0 \\
        \dot{x}(\tau) = u\bigl(x(\tau), \tau\bigr)
    \end{cases}
    \qquad \tau \in [0, 1] \quad .
    \label{eq:fm-ode}
\end{equation}

whose solution induces a trajectory $x(\tau)$ starting from $p_0$. For every $\tau$ in [0, 1], the intermediate marginals of $x(\tau)$ trace a probability path $p(\cdot \mid \tau)$ that continuously deforms $p_0$ into $p_1$. Once $u$ is learned, sampling reduces to drawing $x_0 \sim p_0$ and numerically integrating Equation~\eqref{eq:fm-ode} (e.g., with the Euler method).

%\vspace{0.25em}
\noindent
\paragraph{Conditional Flow Matching.} Rather than learning $u$ directly, conditional flow matching (CFM) conditions on a target sample $x_1 \sim p_1$ (independent of $\tau$), and introduces a conditional probability path
\[
%p(\cdot \mid z = x_1, \tau) = \mathcal{N}\bigl(\tau x_1,\; (1 - \tau)^2 I_d\bigr).
p(\cdot \mid x_1) = \mathcal{N}\bigl(\tau x_1,\; (1 - \tau)^2 I_d\bigr), \qquad x_\tau = \tau x_1 + (1 - \tau)\epsilon, \qquad \epsilon \sim \mathcal{N}\bigl(0, I_d\bigr)
\]
As $\tau$ moves from $0$ to $1$, the mean $\mathbb{E}[x_\tau \!\mid\! x_1] = \tau x_1$ drifts linearly from the origin to $x_1$ while the covariance $(1 - \tau)^2 I_d$ contracts to zero, so that the path collapses to the point mass $\delta_{x_1}$ at $\tau = 1$. However, other choices of conditional path are
possible and we refer the reader to~\citep{lipman2024, gagneux2025, gao2025}. Given the continuity equation~\citep[Section~3.5]{lipman2024}, this construction induces a conditional velocity field with the closed-form expression, whose conditional expectation recovers the optimal velocity field $u^\star(x, \tau)$~\citep[Theorem~1]{lipman2023}:
\begin{equation}
u^{\mathrm{cond}}(x, x_1, \tau)
\;=\; \frac{x_1 - x}{1 - \tau} \quad.
\label{eq:cond-velocity}
\end{equation}
%Marginalizing over $z$ recovers the desired path $p(\cdot \mid \tau)$. The corresponding optimal velocity field is given by~\citep[Thm.~1]{lipman2023}
%\begin{equation}
%v^\star(x, \tau)
%\;=\; \mathbb{E}_{z \mid x, \tau}\!\left[ v^{\mathrm{cond}}(x, z, \tau) \right].
%\label{eq:optimal-velocity}
%\end{equation}
%\vspace{0.25em}
\noindent
\paragraph{Training Objective.} The optimal velocity can be approximated by a neural network ${u_{\theta} : \mathbb{R}^d \times [0,1] \rightarrow \mathbb{R}^d}$ with parameters $\theta$ minimizing
\begin{equation}
\mathcal{L}_{\mathrm{FM}}(\theta)
\;=\;
\mathbb{E}_{\substack{\tau \sim \mathcal{U}([0,1]) \\ x_{\tau} \sim p(\cdot|\tau)}}
\Bigl\| u_\theta(x_\tau, \tau)
-  u^\star(x_{\tau}, \tau)
\Bigr\|^2 \quad .
\label{eq:fm-loss}
\end{equation}
A direct regression onto $u^\star$ would require evaluating the conditional expectation in Equation~\eqref{eq:cond-velocity}, which is generally intractable. The key observation of~\citet{lipman2023} is that regressing against $u^\star$ is equivalent, up to an additive constant, to minimizing
\begin{equation}
\mathcal{L}_{\mathrm{CFM}}(\theta)
\;=\;
\mathbb{E}_{\substack{\tau \sim \mathcal{U}([0,1]) \\ x_0 \sim p_0,\, x_1 \sim p_1}}
\left[
\Bigl\| u_\theta(x_\tau, \tau)
- \underbrace{\frac{x_1 - x_\tau}{1 - \tau}}_{=x_1-x_0}
\Bigr\|^2
\right], \qquad x_\tau := \tau x_1 + (1 - \tau) x_0\quad .
\label{eq:cfm-loss}
\end{equation}
This objective is straightforward to estimate: $\tau$ and $x_0$ are sampled directly, and $p_1$ is replaced by the empirical distribution over the training set. When pairs $(x_0, x_1)$ are drawn from an optimal-transport coupling between $p_0$ and $p_1$, the same linear interpolant $x_\tau$ defines the displacement interpolation underlying OT-CFM \citep{tong2024}. Under the independent sampling $x_0 \sim p_0$, $x_1 \sim p_1$ used above, the objective remains a valid CFM loss but the coupling is not optimal.

\subsection{Discrete Wavelet Transform}\label{sec:dwt}
The DWT decomposes a signal into components simultaneously localized in space (or time) and scale \citep{mallat1989, daubechies1992}. Formally, for a one-dimensional signal $f \in L^2(\mathbb{R})$, a family of discrete wavelets is generated from a single function $\psi \in L^2(\mathbb{R})$, called the \emph{mother wavelet}, through dyadic dilations and discrete translations:
\begin{equation}
    \psi_{j,n}(x) = 2^{-j/2}\,\psi\!\left(2^{-j}x - n\right), 
    \quad (j,n)\in\mathbb{Z}^2 \quad .
\end{equation}
where $j$ controls the scale (fine-to-coarse) and $n$ the spatial location. For suitably constructed $\psi$, the family $\{\psi_{j,n}\}_{j,n}$ forms an \emph{orthonormal basis} of $L^2(\mathbb{R})$, so that $f$ admits the unique multi-scale expansion: 
\begin{equation} \label{eq:dwt}
    f \;=\; \sum_{n} a_J[n]\,\phi_{J,n}
         + \sum_{j=1}^{J}\sum_{n} d_j[n]\,\psi_{j,n} \quad .
\end{equation}
where $a_J[n]$ are coarse-scale \emph{approximation coefficients}, $d_j[n]$ are \emph{detail coefficients} at scale $2^j$, and $\phi$ is the scaling function associated to the mother wavelet~\citep{mallat1989, ha2021, accarino2025}. For a signal $x \in \mathbb{R}^{C \times H \times W}$, where $C$ is the number of channels (or physical variables) and $H \times W$ is the spatial grid, the 2-dimensional DWT is applied separately along rows and columns. We indicate the DWT of the signal $x$ across scales $j \in \{1, \ldots, J\}$, induced by a mother wavelet $\psi$ as:
\begin{equation}
    w_j \;=\; \mathcal{W}_{j}(x) \;\in\; \mathbb{R}^{C \times 4 \times H_j \times W_j},
    \qquad H_j \approx \frac{H}{2^{j}},\quad
    W_j \approx \frac{W}{2^{j}}.
\end{equation}
At each scale, the DWT produces four coefficient maps (or sub-bands), one approximation (LL) and three oriented detail matrices (LH (horizontal), HL (vertical), HH (diagonal)), computed on a progressively down-sampled grid of size $H_j \times W_j$, where $j=1$ corresponds to the finest resolution. The full multi-scale representation is the collection $w = \mathcal{W}(x) = \{w_j\}_{j=1}^{J}$, and exact reconstruction is guaranteed by the inverse DWT (IDWT) by $x = \mathcal{W}^{-1}(w)$. Different mother wavelets exist, and in this work we consider four mother wavelets, $\mathrm{db}1$ (haar), $\mathrm{db}2$, $\mathrm{db}4$, and $\mathrm{db}6$, with filter lengths 2, 4, 8, and 12, respectively (refer to Appendix~\ref{si-sec:wavelets} and Table~\ref{tab:wavelets} for additional details).
%\hl{if there is space add what we need for the IDWT. LL, LH, HL and HH at scale J and (LH, HL, HH) at scales j in [1, J-1]}

\section{Wavelet Flow Matching for Physics Emulation}\label{sec:fm_wavelet}
Rather than learning the conditional distribution $p(x^{t+1}\!\mid\!x^t)$ directly in pixel space, \textbf{Wavelet Flow Matching} (WFM) re-frames the OT-CFM problem  (Section~\ref{sec:ot-cfm}) in the wavelet domain and learns $p(w^{t+1}\!\mid\!w^t)$, where $w^{(\cdot)} = \mathcal{W}(x^{(\cdot)})$ denotes the multi-scale wavelet representation of the state of the system at a specific point in time $(\cdot)$ (Section~\ref{sec:dwt}). This yields $J$ scale-specific flows, sharing one interpolation time $\tau$ and one velocity network, that together define the generative task over the full wavelet representation of the target field. The illustration of the wavelet-space emulation process is depicted in Figure~\ref{fig:graphical_abstract}.

\paragraph{Problem statement.} Let $x^t \in \mathbb{R}^{C \times H \times W}$ denote the state of the system at time $t$. The goal is to learn the conditional distribution $p(w^{t+1} \mid w^t, \{x^{t-l}\}_{l=1}^{L}, \kappa)$, where $\{x^{t-l}\}_{l=1}^{L}$ is a context window of $L$ past states and $\kappa$ is a set of scalar variables encoding static physical properties of the system. We distinguish throughout between physical time $t$ and the OT-CFM interpolation time $\tau \in [0, 1]$.

\paragraph{Multi-scale Optimal Transport Paths.} Rather than constructing a single flow in pixel space, we define $J$ flows, one per wavelet scale, each transporting scale-matched Gaussian noise to the corresponding four wavelet sub-bands (see Section~\ref{sec:dwt}) of the target state:
\begin{equation}
    w_{j}^{\tau} \;=\; \tau\, w_{j}^{t+1} \;+\; (1-\tau)\, \varepsilon_{j},
    \quad \tau \sim \mathcal{U}([0, 1]),
    \quad \varepsilon_j \sim \mathcal{N}(0, I_{d_j}),
    \label{eq:wavelet_interpolant}
\end{equation}
%\begin{equation}
%    w_{j}^{\tau} \;=\; \tau\, w_{j}^{t+1} \;+\; (1-\tau)\, \varepsilon_{j},
%    %\qquad \varepsilon_{j} \sim \mathcal{N}(0, I_d),
%    \quad \tau \sim \mathcal{U}\bigl([0, 1]\bigr),
%    \label{eq:wavelet_interpolant}
%\end{equation}
where $w_j^{t+1} = \mathcal{W}_j(x^{t+1}) \in \mathbb{R}^{C \times 4 \times H_j \times W_j}$
are the wavelet coefficients of the target state at scale $j$, and $\varepsilon_j$ is Gaussian noise drawn independently at each scale. This is merely a practical design choice: since the wavelet families used in this work are orthonormal, applying the DWT to Gaussian noise sampled in pixel space yields independent standard Gaussian coefficients at each scale, so the two sampling procedures are equivalent in distribution, $\mathcal{W}_j(\varepsilon) \overset{d}{=} \varepsilon_j$ \citep{simoncelli1996, donoho1994}. In both cases, the condition on the source distribution in OT-CFM~\citep{lipman2023, lipman2024, tong2024} is satisfied. Therefore, we draw $\varepsilon_j$ directly in wavelet space for sake of simplicity and to avoid additional DWT for the noise.

\paragraph{Velocity-predicting Neural Network.} The velocity network $u_\theta$ is a U-Net~\citep{ronneberger2015} whose encoder--decoder hierarchy naturally mirrors the multi-scale structure of the DWT (see Figure~\ref{fig:unet_diagram}). At each scale $j$, it receives as input the channel-wise concatenation of the wavelet coefficients of the current state $w_j^t = \mathcal{W}_j(x^t)$ and the noisy interpolant $w_j^\tau$, obtained by interpolating between Gaussian noise and the wavelet coefficients of the subsequent state:
\begin{equation}
    z_j^{\tau} \;=\; \bigl[\, w_j^{t},\;\; w_j^{\tau} \,\bigr]_C
    \;\in\; \mathbb{R}^{2C \times 4 \times H_j \times W_j},
    \label{eq:cond-cat}
\end{equation}
where the current-state coefficients $w_j^{t}$ serve as a static, scale-consistent conditioning signal throughout integration. Note that this concatenation does not violate the FM framework: at $\tau = 0$ the interpolated coefficients $w_j^{\tau}$ coincide with pure Gaussian noise $\varepsilon_j$, meaning the model always begins from an unstructured noise field at sampling time. The concatenated wavelet coefficients $z_j^\tau$ at each scale $j$ are injected directly into the corresponding encoder level via a dedicated stem convolution, so that each resolution stage of the U-Net processes the frequency sub-bands it is spatially matched to, as illustrated in Figure~\ref{fig:unet_diagram}. Starting from a shared backbone, the decoder employs $J$ prediction heads, one per scale, each producing the corresponding velocity estimate $\hat{u}_j$ at the appropriate spatial resolution and for all the four sub-bands. Beyond the input in Equation~\eqref{eq:cond-cat}, the U-Net is conditioned on three sources of information, fused into a single vector $c \in \mathbb{R}^{d}$ and injected into every residual block~\citep{he2016} via adaptive group normalization (AdaGN)~\citep{dhariwal2021}, a feature-wise scale and shift~\citep{perez2018} applied after group normalization: \emph{(i)} the FM interpolation time $\tau$ embedded via a sinusoidal Fourier encoding followed by a small MLP; \emph{(ii)} the static physical parameters $\kappa$, embedded linearly; and \emph{(iii)} the temporal context $\{x^{t-l}\}_{l=1}^{L}$, embedded via cross-attention over the $L$ frames prior to $x^t$. Because every head $\hat{u}_j$ is computed from the shared encoder--decoder, it depends on the interpolants at all scales, fine-to-coarse through the encoder injections and coarse-to-fine through the decoder, so the $J$ flows are coupled at every integration step (see Appendix~\ref{si-sec:cross_scale}).

\paragraph{Formulation of Training Objective in Wavelet Space.} As introduced in Section~\ref{sec:ot-cfm} Equation~\eqref{eq:cond-velocity}, the target velocity at scale $j$ along the straight path connecting $\varepsilon_j$ to $w_j^{t+1}$ is constant and is obtained by differentiating the interpolant in Equation~\eqref{eq:wavelet_interpolant} with respect to $\tau$:
\begin{equation}
    u_j \;=\; \frac{d\,w_j^{\tau}}{d\tau}  \;=\; w_j^{t+1} - \varepsilon_j, \qquad u_j \in \mathbb{R}^{C \times 4 \times H_j \times W_j}.
    \label{eq:target_velocity}
\end{equation}
We train a velocity-predicting neural network $u_{\theta}$ to match Equation~\eqref{eq:target_velocity} at every scale simultaneously, where $\hat{u}_j$ denotes the network prediction at scale $j$. To account for the large differences in energy across scales, we extend Equation~\eqref{eq:cfm-loss} by normalizing the per-scale loss by the spatial variance of the target velocity:
\begin{equation}
    \ell_j(\theta) \;=\;
    \mathbb{E}_{\substack{\tau \sim \mathcal{U}[0,\, 1] \\
              (x^t, x^{t+1}) \sim q,\; \varepsilon_j \sim \mathcal{N}(0,I_{d_{j}})}}
    \left[
      \frac{1}{BC}
      \sum_{b,c}
      \frac{
        \bigl\langle \bigl(\hat{u}_{j} - u_{j} \bigr)^2 \bigr\rangle
      }{
        \bigl\langle \bigl(u_{j} - \langle u_{j} \rangle\bigr)^2 
        \bigr\rangle + \epsilon
      }
    \right],
    \label{eq:scale_loss}
\end{equation}
where $q$ denotes the data distribution, $B$ and $C$ are the batch size and number of 
channels respectively, $\langle \cdot \rangle$ denotes spatial averaging over the $H_j \times W_j$ grid, and $\epsilon = 10^{-4}$ is a small constant for numerical stability. The denominator is the per-sample, per-channel spatial variance of the target velocity. This normalization ensures that all scales contribute equally to the gradient, preventing the energy-dominant LL sub-band from overwhelming the loss contribution of the detail sub-bands. Variance-normalized objectives of this kind are increasingly common in physics emulation~\citep{rozet2025}; in wavelet space the normalization is essential, because approximation and detail coefficients differ in energy by orders of magnitude. The total training objective is then a weighted sum across all $J$ scales:
\begin{equation}
    \mathcal{L}(\theta) \;=\; 
    %\sum_{j=0}^{J-1} \bar{\lambda}_j \, \ell_j(\theta),
    \sum_{j=1}^{J} \frac{\lambda_j}{\sum_{j} \lambda_{j}} \, \ell_j(\theta),
    \qquad
    %\bar{\lambda}_j \;=\; \frac{\lambda_j}{\sum_{j'=0}^{J-1} \lambda_{j'}},
    \label{eq:total_loss}
\end{equation}
In all experiments we use uniform weighting, setting $\lambda_j = 1$ for all $j$, so that each scale contributes equally to $\mathcal{L}(\theta)$. 

\paragraph{Inference and Autoregressive Rollout.}
At inference time, sampling is performed entirely in the wavelet space. Given the current physical state $x^t$, we first compute the multi-scale wavelet representation $\mathcal{W}(x^t)$ and independently sample a noise field at each scale $\varepsilon_j \sim \mathcal{N}(0, I_{d_{j}})$ directly in wavelet space. Starting from $w_j^{\tau=0} = \varepsilon_j$ at each scale, the predicted wavelet coefficients of the next state are obtained by integrating the learned velocity field via an Euler 
solver with uniform step size $\Delta\tau = 1/N$ over $\tau \in [0, 1]$:
\begin{equation}
    w_j^{\tau_{n+1}} \;=\; w_j^{\tau_n} \;+\; 
    \Delta\tau \cdot u_\theta\!\left(\bigl[w_j^{t},\, w_j^{\tau_n}\bigr]_C,\; 
    \tau_n,\; \kappa,\; \{x^{t-l}\}_{l=1}^{L}\right),
    \qquad n = 0, \ldots, N-1,
    \label{eq:sampling}
\end{equation}
where at each step the network $u_\theta$ produces multi-scale velocity estimates that are scaled by $\Delta\tau$ and accumulated into the current wavelet state. After $N$ steps, $w_j^{\tau_{N}} = \hat{w}_j^{t+1}$ gives the multi-scale wavelet representation of the next state at physical time $t+1$. The corresponding pixel space state is then recovered in a single call to the IDWT: 
$\hat{x}^{t+1} = \mathcal{W}^{-1}\!\bigl(\{\hat{w}_j^{t+1}\}_{j=1}^{J}\bigr)$ (Section~\ref{sec:dwt}). 
For autoregressive rollout, $\hat{x}^{t+1}$ is fed back as the new current state, its wavelet coefficients $\{\mathcal{W}_j(\hat{x}^{t+1})\}$ are recomputed via the DWT, the context window is updated, and the procedure repeated across many steps.

%$\{w_j^{t+1}\}$ are recomputed via the DWT, the context window is updated, and the procedure repeated across many steps.

\section{Results}
\subsection{Experimental Setup}
In this section, we empirically evaluate \textbf{WFM}'s performance and its ability to accurately emulate complex physical systems across a range of dynamical regimes. We compare WFM against a representation-based suite of baselines. The only exception is FM$_{pixel}$, which operates directly in pixel space and therefore serves as a control reference of generation quality in the original data space. Unless otherwise stated, generative models produce $\mathcal{M}=8$ trajectories from the same initial condition, each drawing fresh Gaussian noise at every autoregressive step, and all scores are averaged over three inference seeds (42, 84, 168) shared across models; deterministic baselines produce a single prediction. Each table caption reports the seed-to-seed variability.

\begin{figure}[t!]
  \centering
  \begin{subfigure}{1.0\textwidth}
    \includegraphics[width=\linewidth]{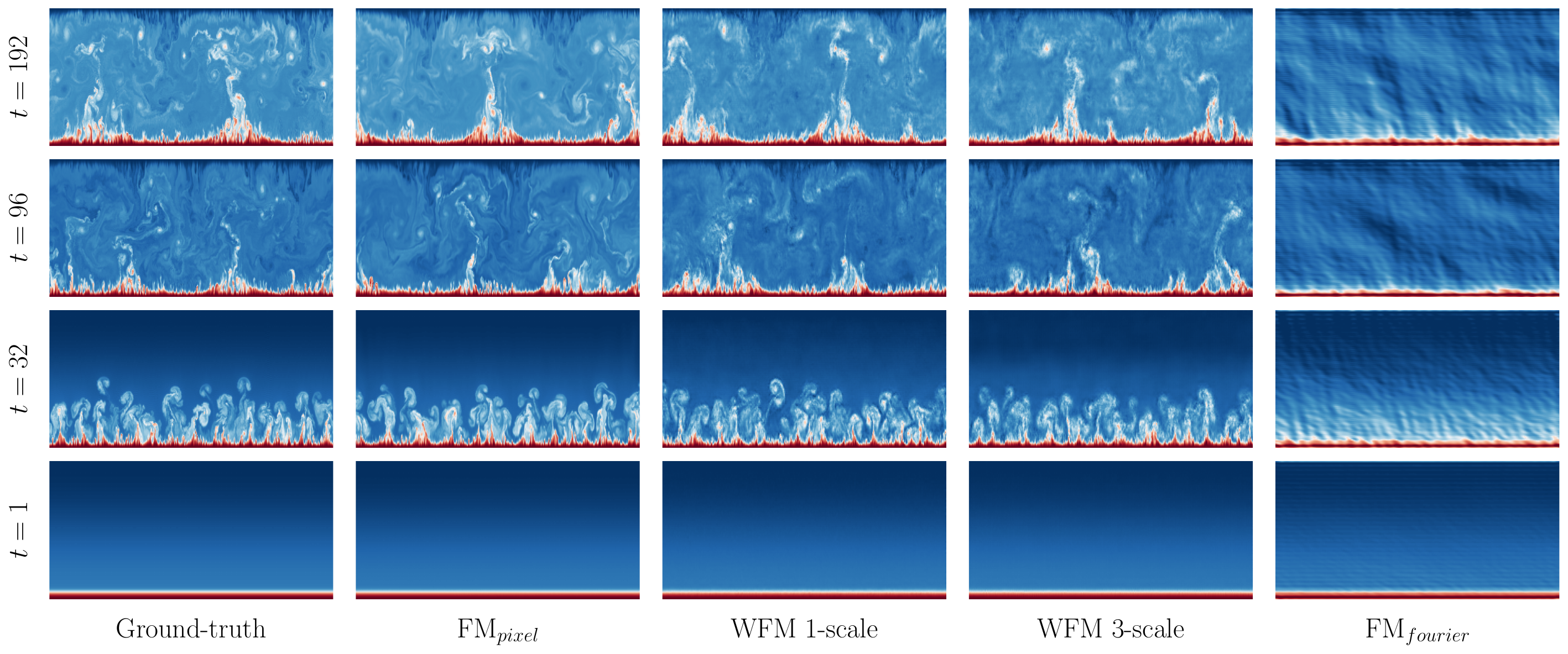}
    \caption{Rayleigh--Bénard Convection}
    \label{fig:qual_rb}
  \end{subfigure}
  \begin{subfigure}{1.0\textwidth}
    \includegraphics[width=\linewidth]{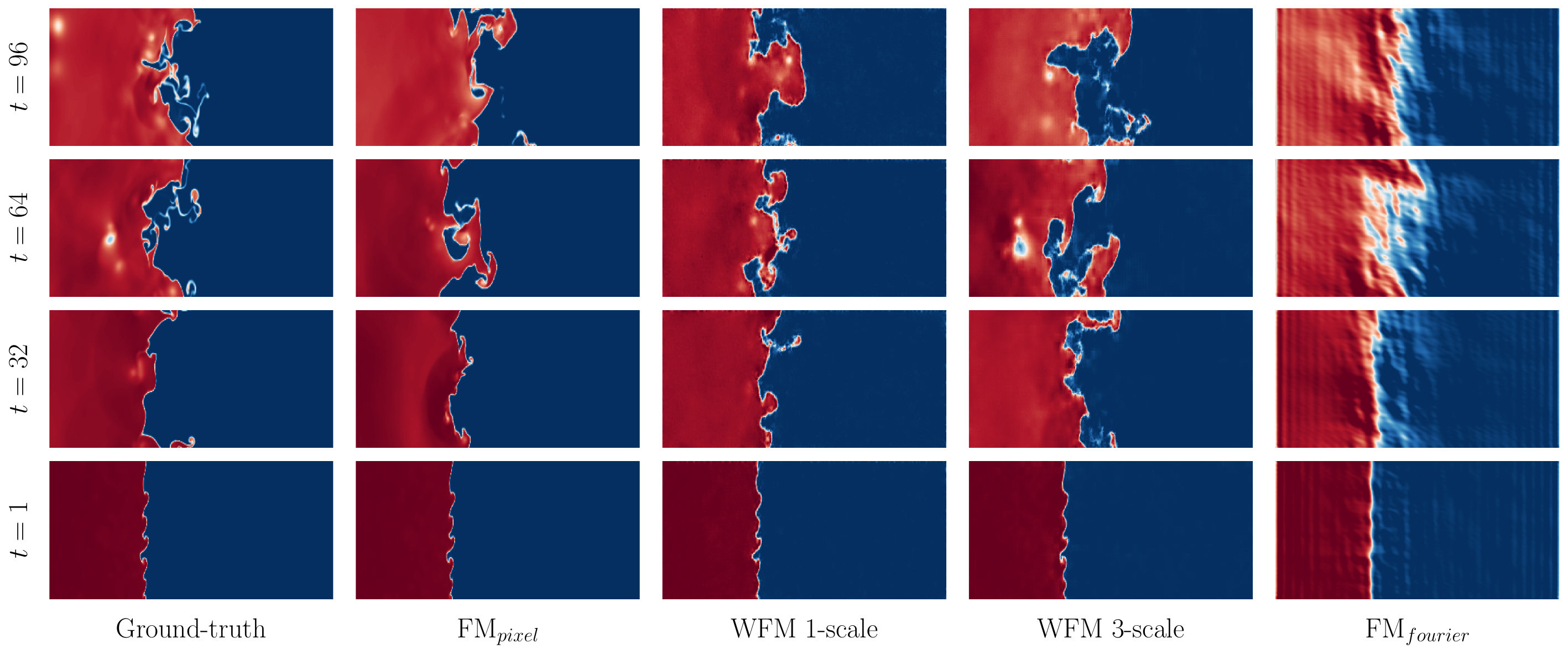}
    \caption{Turbulent Radiative Layer 2D}
    \label{fig:qual_trl}
  \end{subfigure}
  \caption{Qualitative results at different rollout lead-times ($t$).}
  \vspace{-0.5cm}
\end{figure}

\paragraph{Datasets.} To study the effect of the generation across multiple levels of decomposition in wavelet space, we select three PDE datasets with different dynamical characteristics from The Well~\cite{ohana2024}, including Turbulent Radiative Layer 2D (TRL) \citep{fielding2020}, Rayleigh--Bénard (RB) convection \cite{burns2020}, and Active Matter (AM) \citep{maddu2024}. These datasets represent distinct fluid regimes that cover many key challenges in dynamical systems emulation, including non-linearities, multi-scale interactions, and complex spatio-temporal patterns. Additional details are provided in Appendix~\ref{si-sec:datasets}.

\paragraph{WFM variants.} WFM is built on a U-Net encoder--decoder~\cite{ronneberger2015} that operates across wavelet scales (Section~\ref{sec:fm_wavelet}). To isolate the effect of multi-scale generation, we train both a 1-scale $(J=1)$ and 3-scale WFM $(J=3)$ for each PDE system. We set the maximum number of wavelet scales to $J=3$, as this is the largest value compatible with all grid resolutions considered. For grids with a minimum spatial dimension of 128 (i.e., TRL and RB), the coarsest approximation sub-band at $J=3$ has size $(\cdot)\times 16$; increasing to $J=4$ would reduce the bottleneck dimension in the U-Net to 8, which is insufficient for meaningful convolutional feature extraction. WFM conditions on $L=3$ past frames, alongside the current state $x^{t}$, yielding a total temporal context of four states, consistent with baseline models in The Well \citep{ohana2024}. Past frames are provided in pixel space, as they serve only to inform the temporal context and need not be decomposed into frequency sub-bands. We fix the embedding dimension to $d = 256$ throughout all experiments. Additional information is provided in Appendix~\ref{si-sec:baselines}.

\paragraph{Baselines.} To assess the quality of WFM's wavelet-space generation, we compare against flow matching in pixel space (FM$_{pixel}$), flow matching in Fourier space (FM$_{fourier}$) as described in~\cite{wang2025fourierflow}, Wavelet Diffusion Neural Operator (WDNO)~\cite{hu2024wdno}, Fourier Neural Operator (FNO)~\citep{li2020fno}, a Tensorized FNO (TFNO)~\cite{kossaifi2023tfno}, and a Wavelet Neural Operator (WNO)~\cite{tripura2023wno}, which represent the main families of generative and deterministic PDE surrogates. All flow-matching models use the same U-Net architecture, flow-matching objective and sampling budget ($N=50$ Euler steps, $\mathcal{M}=8$ trajectories rolled out from the same initial condition); apart from learning rate and batch size, they differ only in the space where generation is performed. Capacity is matched pairwise: WFM 1-scale and FM$_{pixel}$ are architecturally identical (12.7M parameters), and WFM 3-scale and FM$_{fourier}$ have the same capacity ($\approx$30.5M parameters; Appendix~\ref{si-sec:baselines}). Differences within each pair can therefore be attributed to the representation rather than to architecture or parameter count. On Rayleigh--B\'enard we also compare against a latent-space model, LatentFM$_{\div 64}$, which performs flow matching in the latent space of the pre-trained autoencoder of~\citet{rozet2025} (64$\times$ compression) with the same flow-matching objective and sampler as WFM (Appendix~\ref{si-sec:latentfm}). %Additional information is provided in Appendix~\ref{si-sec:baselines}.

\paragraph{Metrics.} To guarantee a comprehensive comparison of WFM across scales we consider multiple metrics. Similar to \cite{ohana2024,rozet2025}, our analysis is either performed at a lead time $t = i \times \Delta$ or averaged across lead times $a{:}b$. Specifically, we include the variance-normalized root mean squared error (VRMSE), which compares better across scales and in low-variance regions; the continuous ranked probability score (CRPS)~\cite{gneiting2007}, which generalizes the mean absolute error to probabilistic forecasts and, computed over $\mathcal{M}$ independent trajectories rolled out from the same initial condition, rewards both accuracy and calibrated spread; and the spectral coherence RMSE~\citep{rozet2025}, which measures the error across frequency bands. Full definitions are provided in Appendix~\ref{si-sec:metrics}.

%To guarantee a comprehensive comparison of WFM across scales we consider multiple metrics. Similar to \cite{ohana2024,rozet2025}, our analysis is either performed at a lead time $\tau = i \times \Delta$ or averaged across lead times $a:b$. Specifically, we include the variance-normalized root mean squared error (VRMSE) as it compares better across scales and for low-variance regions, the continuous ranked probability score (CRPS)~\cite{gneiting2007}, which generalizes the mean absolute error to probabilistic forecasts, and the spectral coherence RMSE~\citep{rozet2025}, to assess the spectral error across frequency bands. Because generative models return $\mathcal{M}$ samples at every step, their spread provides an estimate of predictive uncertainty; CRPS evaluates this uncertainty, rewarding both accuracy and calibrated spread (Appendix~\ref{si-sec:metrics}). Full definitions of these metrics are provided in Appendix~\ref{si-sec:metrics}.

\begin{table}[ht!]
\centering
\caption{Spectral Coherence RMSE averaged across \textit{Low}, \textit{Mid} and \textit{High} frequency bands and rollout horizons per PDE system. For each WFM variant, \textbf{bold} denotes the optimal wavelet representation used in all subsequent experiments. LatentFM$_{\div 64}$ is evaluated on Rayleigh--B\'enard only. Mean over three inference seeds (42, 84, 168); the standard deviation across seeds is below 5\% of the mean for all entries. Here, -- indicates that the model was not available for the corresponding system.}
\label{tab:avg_sc_rmse_gen_wavelets}
\begin{tabular}{lccc}
\toprule
 & Turbulent Radiative Layer 2D & Rayleigh--B\'{e}nard & Active Matter \\
\midrule
WFM-1 haar & \textbf{0.147} & 0.140 & \textbf{0.241} \\
WFM-1 db2 & 0.154 & 0.132 & 0.288 \\
WFM-1 db4 & 0.149 & \textbf{0.121} & 0.303 \\
WFM-1 db6 & 0.150 & 0.140 & 0.290 \\
\midrule
WFM-3 haar & \textbf{0.149} & 0.131 & 0.247 \\
WFM-3 db2 & 0.154 & 0.111 & \textbf{0.220} \\
WFM-3 db4 & 0.154 & \textbf{0.109} & 0.249 \\
WFM-3 db6 & 0.153 & 0.112 & 0.238 \\
\midrule
FM$_\text{pixel}$ & 0.153 & 0.131 & 0.244 \\
FM$_\text{fourier}$ & 0.379 & 0.266 & 0.376 \\
WDNO & 0.271 & 0.467 & 0.304 \\
LatentFM$_{\div 64}$ & -- & 0.162 & -- \\
\bottomrule
\end{tabular}
\end{table}

\subsection{WFM Evaluation}\label{sec:wfm_eval}
%We benchmark WFM variants (1- and 3-scale) across four prescribed mother wavelets, haar, $\mathrm{db}2$, $\mathrm{db}4$, and $\mathrm{db}6$, evaluated independently for each PDE system, and compare them to other generative models. To select the optimal wavelet representation, we report the spectral coherence RMSE averaged across \textit{Low}, \textit{Mid}, and \textit{High} frequency bands (Table~\ref{tab:avg_sc_rmse_gen_wavelets}), which provides a holistic measure of spectral fidelity across the full frequency spectrum. Per-band results are further detailed in Table~\ref{tab:appendix_spectral}.
We benchmark WFM 1- and 3-scale with four prescribed mother wavelets (haar, $\mathrm{db}2$, $\mathrm{db}4$, $\mathrm{db}6$) on each PDE system and compare them to other generative models. The wavelet is selected by a single criterion, the spectral coherence RMSE averaged over the \textit{Low}, \textit{Mid} and \textit{High} bands on the test trajectories (Table~\ref{tab:avg_sc_rmse_gen_wavelets}), which provides a holistic measure of spectral fidelity across the full frequency spectrum. Tables~\ref{si-tab:appendix_vrmse}--\ref{tab:appendix_spectral} report every wavelet on all metrics, including per-band results, so the choice can be checked against VRMSE and CRPS.

Both WFM 1- and 3-scale consistently outperform all generative baselines across every PDE system and frequency band. For WFM 1-scale, $\mathrm{haar}$ achieves the lowest average spectral coherence RMSE on TRL and AM, while $\mathrm{db}4$ is optimal for RB. For WFM 3-scale, $\mathrm{haar}$ yields the best representation on TRL, $\mathrm{db}4$ on RB, and $\mathrm{db}2$ on AM. All subsequent experiments adopt these PDE-specific wavelet selections, which we summarize in Table~\ref{si-tab:optimal_wavelets}.

Qualitative results are presented in Figures~\ref{fig:qual_rb}--\ref{fig:qual_trl}. Both WFM variants produce physically plausible states at extended rollout steps, remaining comparable to pixel-space generation. The difference between variants is most evident on AM (Figure~\ref{si-fig:qual_am}), where WFM 1-scale loses coherent structure as early as $t=16$, whereas WFM 3-scale preserves it throughout.

\begin{figure}[ht!]
    \centering
    \begin{subfigure}{\textwidth}
        \centering
        \caption{VRMSE for WFM variants and deterministic baselines.}
        \includegraphics[width=\textwidth]{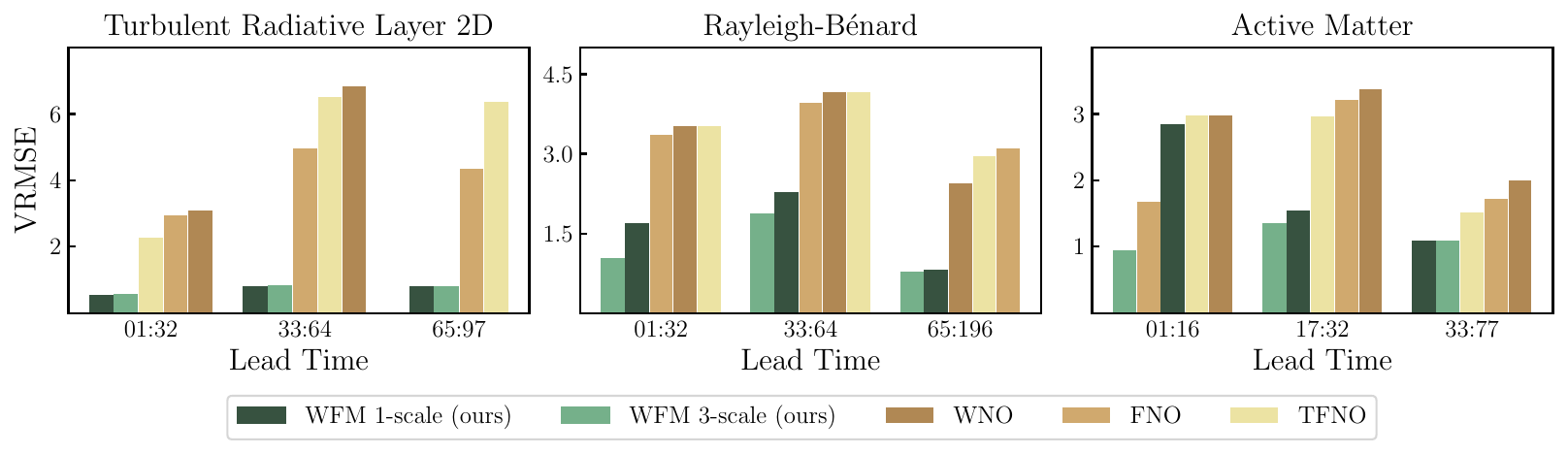}
        \label{fig:vrmse_det}
    \end{subfigure}
    \hfill
    \begin{subfigure}{\textwidth}
        \centering
        \caption{CRPS for WFM variants and generative baselines.}
        \includegraphics[width=\textwidth]{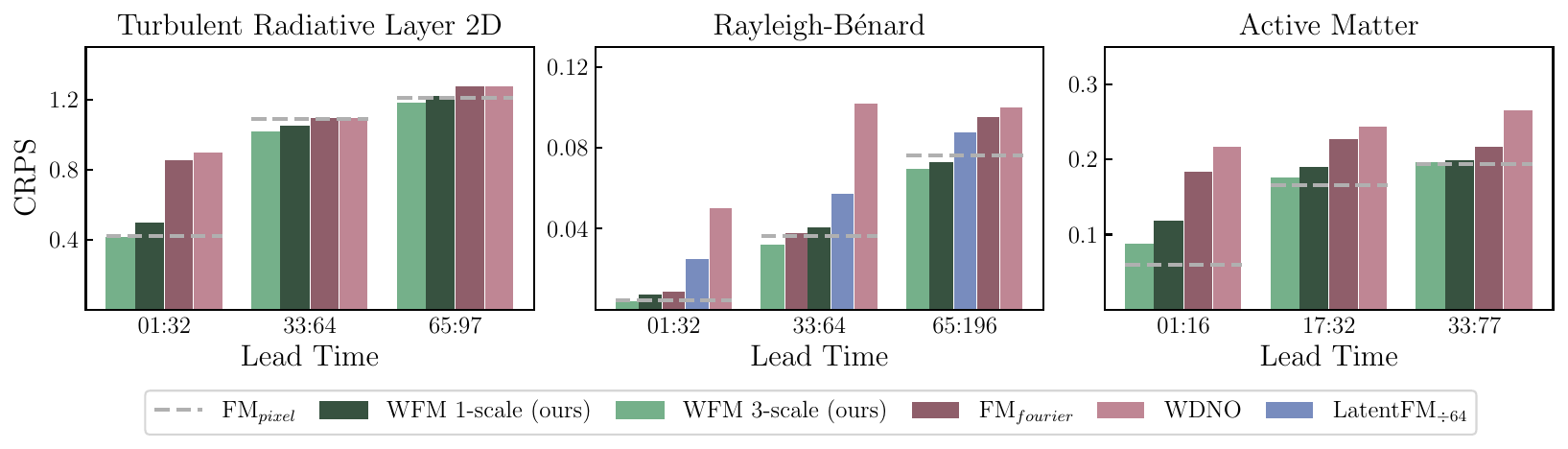}
        \label{fig:crps_gen}
    \end{subfigure}
    \hfill
    \begin{subfigure}{\textwidth}
        \centering
        \caption{Spectral Coherence RMSE for WFM variants and generative baselines.}
        \includegraphics[width=\textwidth]{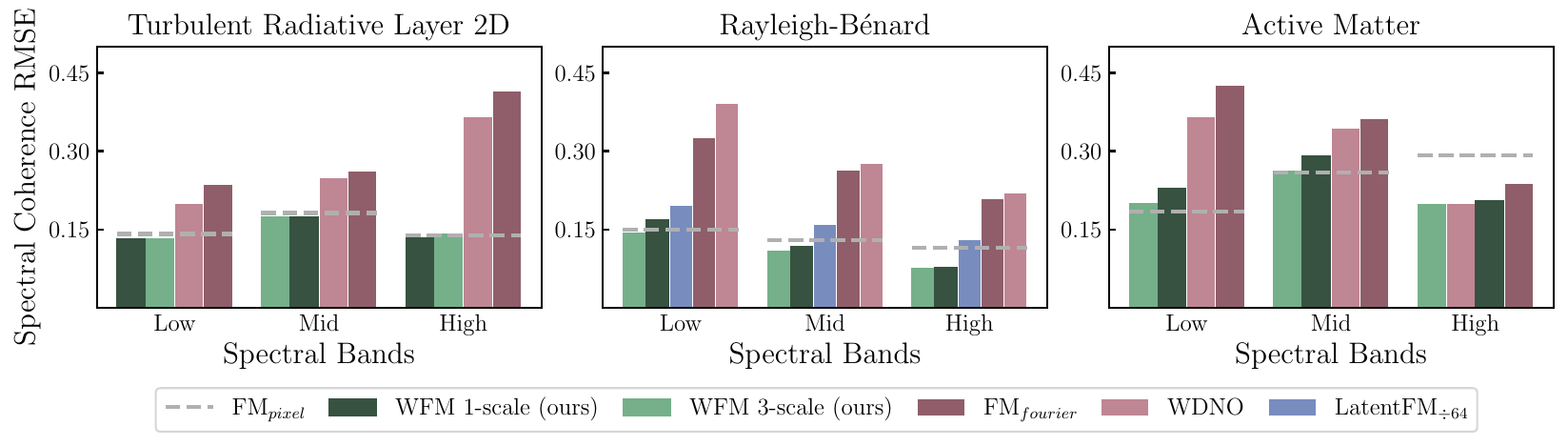}
        \label{fig:sc_gen}
    \end{subfigure}
    \caption{VRMSE (a) and CRPS (b) averaged over rollout horizons, and Spectral coherence RMSE (c) per frequency band, for the three PDE systems. Generative models are averaged over three inference seeds; FNO, TFNO and WNO are deterministic. FM$_{pixel}$ is shown as a dashed gray line and serves as a quality-control reference.}
    \label{fig:metrics}
\end{figure}

WFM performance relative to deterministic and generative baselines is shown in Figure~\ref{fig:metrics}. WFM 3-scale outperforms all deterministic baselines (Figure~\ref{fig:vrmse_det}) and the other generative baselines (Figures~\ref{fig:crps_gen} and~\ref{fig:sc_gen}) across all systems. It is also competitive with the quality-control model FM$_{pixel}$, achieving a lower spectral coherence RMSE on RB in all frequency bands and on AM in the high-frequency band.

\begin{wraptable}[8]{r}{0.42\textwidth}
\vspace{-\baselineskip}
\caption{Inference speedup of WFM variants relative to FM$_{pixel}$ (higher is better).}
\label{tab:inference_speedup}
\centering\small
\begin{tabular}{lcc}
\toprule
\textbf{PDE} & \textbf{WFM 1-scale} & \textbf{WFM 3-scale} \\
\midrule
TRL & 1.25$\times$ & 1.14$\times$ \\
RB  & 1.52$\times$ & 1.38$\times$ \\
AM  & 1.46$\times$ & 1.26$\times$ \\
\bottomrule
\end{tabular}
\end{wraptable}

Beyond competitive accuracy, WFM consistently delivers faster inference than FM$_{pixel}$ across all three systems, as reported in Table~\ref{tab:inference_speedup}. Since all experiments are conducted under identical conditions, we attribute this result to the spatially smaller wavelet representation space in which the generative process operates. 
We report the inference profiling results for the selected WFM variants in Table~\ref{tab:inference_perf_ablation_selected}, and provide a comprehensive profiling analysis covering all considered mother wavelets in Table~\ref{tab:inference_perf_ablation}.

%We report additional profiling information for all WFM variants (scales and choice of mother wavelet) in Table~\ref{tab:inference_perf_ablation}. \ref{tab:inference_perf_selected}

%We first report the inference performance of the selected WFM configuration in Table~\ref{tab:inference_perf_selected}, followed by an extensive profiling and ablation analysis of all WFM variants, including different scales and mother wavelet choices, in Table~\ref{tab:inference_perf_ablation}.

%We provide additional evaluation of WFM's performance across different turbulent regimes. For TRL, we vary the cooling time $t_{\mathrm{cool}}$ and report CRPS (Tables~\ref{tab:ablation_trl_crps_tcool_main} and~\ref{tab:ablation_trl_crps_tcool}) and spectral coherence RMSE (Table~\ref{tab:ablation_trl_spectral_tcool}). WFM outperforms all generative baselines across the full range of $t_{\mathrm{cool}}$, with the largest gains in the small $t_{\mathrm{cool}}$ regime, where rapid condensation produces a highly corrugated interface and sharp density contrasts that are particularly challenging to emulate. For RB, we vary $Ra$ and $Pr$ numbers and report CRPS (Tables~\ref{tab:ablation_rb_crps_ra_main},~\ref{tab:ablation_rb_crps_ra}, and ~\ref{tab:ablation_rb_crps_pr}) and spectral coherence RMSE (Tables~\ref{tab:ablation_rb_spectral_ra} and \ref{tab:ablation_rb_spectral_pr}). Even for this PDE, WFM outperforms other models in the full $Ra$--$Pr$ parameter space, including the most challenging regime ($Ra=10^{10}$, $Pr=10^{-1}$), where dynamics are mostly chaotic.

We further evaluate WFM across turbulent regimes. For TRL, we vary the cooling time $t_{\mathrm{cool}}$ and report CRPS (Tables~\ref{tab:ablation_trl_crps_tcool_main} and~\ref{tab:ablation_trl_crps_tcool}) and spectral coherence RMSE (Table~\ref{tab:ablation_trl_spectral_tcool}). WFM is most advantageous in the small $t_{\mathrm{cool}}$ regime, where rapid condensation produces a highly corrugated interface and sharp density contrasts that are particularly challenging to emulate. For RB, we vary the $Ra$ and $Pr$ numbers and report CRPS (Tables~\ref{tab:ablation_rb_crps_ra_main},~\ref{tab:ablation_rb_crps_ra} and~\ref{tab:ablation_rb_crps_pr}) and spectral coherence RMSE (Tables~\ref{tab:ablation_rb_spectral_ra} and~\ref{tab:ablation_rb_spectral_pr}). Here, WFM achieves the lowest CRPS across the full $Ra$--$Pr$ parameter space, including the most chaotic regime ($Ra=10^{10}$, $Pr=10^{-1}$).

\paragraph{Cross-scale coupling, physical consistency and robustness.} Although WFM defines one flow per scale, the flows are coupled through the shared U-Net: perturbing one scale changes the velocity predicted at all others, which independent per-scale flows would not (Appendix~\ref{si-sec:cross_scale}). WFM also preserves the kinetic energy of the reference trajectories, ranking first or second at every rollout horizon on all systems (Table~\ref{tab:ke_mae}). Finally, its rollouts are robust to perturbations of the initial conditions, with CRPS changing by at most 2\% (Appendix~\ref{si-sec:ic_pert}).

%Although WFM defines one flow per scale, the flows are coupled through the shared U-Net: on TRL, perturbing the interpolant at one scale changes the velocity predicted at the other scales, with 15.1\% of the total sensitivity off the diagonal, whereas independent per-scale flows would give exactly zero (Appendix~\ref{si-sec:cross_scale}). Beyond statistical scores, WFM 3-scale tracks the kinetic energy of the reference trajectories as well as FM$_{pixel}$ on TRL, and better at the longest horizon (0.147 vs.\ 0.177); on RB and AM its late-horizon error is higher than FM$_{pixel}$'s but below FM$_{fourier}$ and WDNO (\hl{Table}~\ref{tab:ke_mae}). Finally, perturbing the initial conditions of TRL with Gaussian noise of up to 10\% of each channel's standard deviation changes CRPS by at most 2.1\% and leaves spectral coherence unchanged at three decimals (\hl{Appendix}~\ref{si-sec:ic_pert}).

\section{Discussion}\label{sec:discussion}
A key perspective offered by our approach is its connection to multi-scale closure modeling in PDEs. Classical closure schemes parameterize unresolved small-scale processes as functions of resolved variables, often assuming a unidirectional dependence that neglects cross-scale feedbacks such as backscatter and intermittency~\cite{maeyama2020extracting, sanderse2024scientific}. WFM instead models the joint evolution of all scales in a unified generative framework, where multi-scale representations are induced by the choice of mother wavelet. This provides a principled multi-scale decomposition aligned with the locality and cascade structure of many physical systems, without requiring learned compression and loss of information. The resulting separation of representation and dynamics introduces a structured inductive bias that may improve interpretability~\citep{ha2021} and generalization~\cite{nathaniel2025deep}, as suggested by our consistent performance across the wide range of $Ra$--$Pr$ regimes in RB and $t_{\mathrm{cool}}$ times in TRL.

\begin{table}[ht]
\centering
\caption{CRPS averaged over rollout steps, for the three shortest cooling times $t_{\mathrm{cool}}$ in TRL (a) and per Rayleigh number $Ra$, averaged over Prandtl numbers, in RB (b). Mean over three inference seeds (42, 84, 168); the standard deviation across seeds is below 5\% of the mean for all entries except those marked $^{*}$. \textbf{Bold}: best; \underline{underline}: second best.}
%\caption{CRPS averaged over rollout steps, broken down by the most chaotic parameters combination in TRL (a) and RB (b). Mean over three inference seeds (42, 84, 168); the standard deviation across seeds is below 5\% of the mean for all entries except those marked $^{*}$. \textbf{Bold}: best; \underline{underline}: second best.}
\label{tab:ablation_crps}
\begin{subtable}[t]{0.48\linewidth}
    \centering
    \caption{TRL per $t_{\mathrm{cool}}$.}
    \label{tab:ablation_trl_crps_tcool_main}
    \begin{tabular}{lccc}
    \toprule
    $t_{\mathrm{cool}}$ & 0.03 & 0.06 & 0.1 \\
    \midrule
    WFM-1 haar & 1.165 & 1.135 & 0.865$^{*}$ \\
    WFM-1 db2 & 1.112 & 1.078 & 0.878 \\
    WFM-1 db4 & 1.170 & 1.102 & 0.847$^{*}$ \\
    WFM-1 db6 & 1.076 & 1.074 & \underline{0.831} \\
    \midrule
    WFM-3 haar & 1.107 & 1.117 & 0.868 \\
    WFM-3 db2 & 1.248 & 1.141 & 0.894 \\
    WFM-3 db4 & \textbf{1.048} & \textbf{1.006} & \textbf{0.817}$^{*}$ \\
    WFM-3 db6 & \underline{1.075} & \underline{1.034} & 0.882 \\
    \midrule
    FM$_\text{pixel}$ & 1.220 & 1.126 & 0.875$^{*}$ \\
    FM$_\text{fourier}$ & 1.439 & 1.424 & 1.195 \\
    WDNO & 3.330 & 3.292 & 3.183 \\
    \bottomrule
\end{tabular}
\end{subtable}
\hfill
\begin{subtable}[t]{0.48\linewidth}
    \centering
    \caption{Rayleigh--B\'{e}nard per $Ra$, averaged over $Pr$.}
    \label{tab:ablation_rb_crps_ra_main}
    \begin{tabular}{lccc}
    \toprule
    $Ra$ & $10^{6}$ & $10^{8}$ & $10^{10}$ \\
    \midrule
    WFM-1 haar & 0.059 & 0.060$^{*}$ & 0.051 \\
    WFM-1 db2 & \textbf{0.052} & 0.062$^{*}$ & 0.052$^{*}$ \\
    WFM-1 db4 & 0.059$^{*}$ & 0.060 & 0.051 \\
    WFM-1 db6 & 0.060$^{*}$ & 0.069$^{*}$ & 0.075 \\
    \midrule
    WFM-3 haar & 0.060$^{*}$ & 0.060 & 0.055$^{*}$ \\
    WFM-3 db2 & 0.054$^{*}$ & \underline{0.055} & \underline{0.049} \\
    WFM-3 db4 & \underline{0.053}$^{*}$ & 0.058 & \textbf{0.048}$^{*}$ \\
    WFM-3 db6 & 0.065$^{*}$ & \textbf{0.055} & 0.057 \\
    \midrule
    FM$_\text{pixel}$ & 0.057$^{*}$ & 0.066$^{*}$ & 0.051 \\
    FM$_\text{fourier}$ & 0.071 & 0.081 & 0.065 \\
    WDNO & 0.120 & 0.117 & 0.116 \\
    LatentFM$_{\div 64}$ & 0.087$^{*}$ & 0.067 & 0.065 \\
    \bottomrule
    \end{tabular}
\end{subtable}
\end{table}

\paragraph{Related work.} Hybrid representations that combine data-driven and physics-informed constructions, such as operator-theoretic embeddings~\cite{nathaniel2026generative} or Fourier-based transformations~\cite{oommen2024integrating}, still rely on learning and often on empirical mode truncation or handcrafted basis selection, while multi-scale structure is typically enforced indirectly through hierarchical architectures, e.g.\ pyramidal flow matching~\cite{irvin2025spatiotemporal}, rather than by the representation itself~\cite{accarino2025}. Wavelets offer such a representation: analytically grounded, spatially localized, orthogonal and energy-compacting, with no training required. In generative modeling they have served as cascades of per-scale score models~\citep{guth2022wavelet}, tokens for autoregressive image generation~\citep{mattar2024wavelets}, a training signal for latent diffusion~\citep{sigillo2026latent}, and components of adversarial neural operators~\citep{rani2025generative}. In PDE emulation, WDNO~\citep{hu2024wdno} denoises space--time blocks of wavelet coefficients, Fourier flow matching~\citep{wang2025fourierflow} generates a truncated global spectrum, and latent models~\citep{rozet2025} learn a lossy representation that requires pre-training; WFM instead generates all wavelet coefficients of the next state, coupled across scales through a shared network.

%\paragraph{Limitations.} This study focuses on WFM variants based on the Daubechies family of orthonormal wavelets, but other types such as \textit{mallat}~\citep{mallat2009}, \textit{symlets} and \textit{coiflets}~\citep{daubechies1992} remain unexplored. Additionally, the current DWT implementation operates on regular grids. Extending it to other geometries, such as spherical domains, would extend the applicability of the framework to a wider range of physical systems, including the burgeoning field of weather and climate emulation.

%\paragraph{Limitations.} WFM inherits the assumptions of the DWT. It requires a regular grid, so complex geometries and irregular meshes are not supported; spherical wavelets~\citep{price2024differentiable} are a natural extension towards weather and climate emulation. Non-periodic boundaries rely on padding, which can degrade boundary-dominated dynamics, and discontinuities misaligned with the dyadic grid spread over many coefficients. The mother wavelet is selected per dataset, only the Daubechies family was explored (\textit{symlets} and \textit{coiflets}~\citep{daubechies1992} remain untested), and the number of scales is bounded by the grid size. Gains over pixel-space flow matching depend on the system (on AM, FM$_{pixel}$ retains a lower CRPS), the speedup is moderate (1.1--1.5$\times$), and WFM provides no guarantee of physical conservation or of stability beyond the evaluated horizons.

\paragraph{Limitations.} This study considers only the Daubechies family of orthonormal wavelets; other families, such as \textit{symlets} and \textit{coiflets}~\citep{daubechies1992}, remain unexplored, and the best mother wavelet is currently selected per dataset rather than learned or chosen a priori. The DWT also assumes regular grids and handles non-periodic boundaries through padding, which may affect boundary-dominated dynamics; extending WFM to irregular meshes and other geometries, such as spherical domains~\citep{price2024differentiable, donno2026neptune}, would broaden its applicability, including to weather and climate emulation.

\begin{ack}
We acknowledge funding from NSF through the Learning the Earth with Artificial intelligence and Physics (LEAP) Science and Technology Center (STC) (Award \#2019625). VA acknowledges support from a PIVOT Research award (Award \#12871) from the Simons Foundation. This work used Derecho-GPU system at NCAR through allocation CIS251353 from the \href{https://access-ci.org}{Advanced Cyberinfrastructure Coordination Ecosystem}: Services \& Support (ACCESS) program, which is supported by U.S. National Science Foundation grants \#2138259, \#2138286, \#2138307, \#2137603, and \#2138296.
\end{ack}

\clearpage
\newpage

% \section*{References}
\bibliographystyle{unsrtnat}
\bibliography{references}

\clearpage
\newpage

\appendix

% Reset and prefix appendix floats

\setcounter{figure}{0}
\setcounter{table}{0}

\renewcommand{\thefigure}{S\arabic{figure}}
\renewcommand{\thetable}{S\arabic{table}}

\section{Datasets}\label{si-sec:datasets}

For all PDE datasets, each field is standardized using the mean and standard deviation computed over the training set. Each dataset also provides constant scalar parameters (see Table~\ref{tab:pde_characteristics}), denoted $\kappa$, which are linearly embedded and injected into every residual block of the velocity-predicting U-Net via AdaGN conditioning (Appendix~\ref{sec:v_pred_arch}). For the Rayleigh--Bénard dataset, the Rayleigh and Prandtl numbers are log-transformed prior to embedding.

% Ragged-right p column shorthand
\newcolumntype{P}[1]{>{\centering\arraybackslash}p{#1}}

\begin{table}[ht]
\centering
\caption{Characteristics of the PDE systems considered in this study. Additional details can be found in~\citep{ohana2024}.}
\label{tab:pde_characteristics}
\resizebox{\linewidth}{!}{%
\begin{tabular}{l P{3.8cm} P{3.8cm} P{3.8cm}}
\toprule
 & \textbf{Turbulent Radiative Layer 2D}
 & \textbf{Rayleigh--Bénard}
 & \textbf{Active Matter} \\
\midrule
Fields
  & density, pressure, velocities ($x$, $y$)
  & buoyancy, pressure, velocities ($x$, $y$)
  & concentration, velocities ($x$, $y$), $D_{xx}$, $D_{xy}$,
    $D_{yx}$, $D_{yy}$, $E_{xx}$, $E_{xy}$, $E_{yx}$, $E_{yy}$ \\
\addlinespace
Scalar constants $(\kappa)$  & $t_{\mathrm{cool}}$ & $Ra$, $Pr$ & $\alpha$, $\zeta$ \\
\addlinespace
Resolution (pixel)           & $384\times128$ & $512\times128$ & $256\times256$ \\
\addlinespace
\# Channels ($C$)            & 4    & 4    & 11 \\
\addlinespace
\# Trajectories              & 90   & 1750 & 225 \\
\addlinespace
\# Steps per trajectory      & 101  & 200  & 81 \\
\addlinespace
Size (GB)                    & 6.9  & 342  & 51.3 \\
\addlinespace
Software                     & Athena++ & Dedalus & Python \\
\addlinespace
Reference                    & \citep{fielding2020} & \citep{burns2020} & \citep{maddu2024} \\
\bottomrule
\end{tabular}%
}
\end{table}

\paragraph{Turbulent Radiative Layer 2D.} This PDE \citep{fielding2020} simulates two-dimensional radiative turbulent mixing layers, a ubiquitous phenomenon in astrophysical environments such as the circum-galactic medium, where cold dense gas and hot dilute gas move relative to each other at subsonic velocities. The velocity shear drives the Kelvin–Helmholtz instability, populating intermediate-temperature gas through turbulent mixing; this mixed gas cools rapidly, driving a net mass flux from the hot to the cold phase. The evolution is governed by the compressible Euler equations with a
radiative cooling source term,
\begin{equation}
    \frac{\partial \rho}{\partial t} + \nabla \cdot ( \rho v ) = 0
\end{equation}   
\begin{equation}
    \frac{\partial \rho v}{\partial t}
    + \nabla \cdot ( \rho vv + P ) = 0
\end{equation}
\begin{equation}
    \frac{\partial E}{\partial t}
    + \nabla \cdot [( E + P ) v] = -E / t_{\mathrm{cool}}
\end{equation}
\begin{equation}
    E = P/(\gamma-1), \qquad \gamma = 5/3
\end{equation}

with $\rho$ the density, $v$ the 2D velocity, $P$ the pressure, $E$ the total energy, and $t_{\mathrm{cool}}$ is the radiative cooling time. The dataset comprises 90 trajectories spanning nine values of $t_{\mathrm{cool}} = \{0.03,0.06,0.1,0.18,0.32,0.56,1.00,1.78,3.16\}$ (ten random seeds each), simulated on a $384\times128$ uniform Cartesian grid and stored at 101 snapshots. The fields available are density, pressure, and velocity components, providing a challenging testbed for emulators across a wide range of cooling-to-mixing dynamical regimes.

%Radiative turbulent mixing layers arise in astrophysical environments where cold, dense gas interacts with a hotter surrounding medium, such as cold clouds embedded in hot winds, supernova remnants, or the circum-galactic medium. At the interface between the two phases, velocity shear drives the Kelvin–Helmholtz instability, generating turbulence that promotes mixing and creates an intermediate-temperature phase. This mixed gas cools efficiently via radiative losses, allowing it to condense and join the cold phase if cooling proceeds faster than mixing, or conversely leading to evaporation if mixing dominates. The dynamics of the system are governed by a small set of dimensionless parameters, including the ratio of cooling to shear timescales, the density contrast between cold and hot phases, and the relative Mach number of the flow, which together determine whether cold gas grows or is destroyed. In quasi-steady state, radiative cooling is balanced by the advection of hot, high-enthalpy gas into a thin, highly corrugated cooling layer, whose structure is shaped by turbulence and exhibits fractal-like geometry. These processes are inherently small-scale and difficult to resolve, yet they strongly influence large-scale gas evolution. In our setup, we consider two-dimensional simulations of radiative mixing layers \citep{fielding2020} that capture the coupled evolution of density, temperature, and velocity fields, providing a challenging testbed for modelling the interplay between turbulence, mixing, and cooling across a range of dynamical regimes. \hl{TBC}

\paragraph{Rayleigh--Bénard Convection.} Rayleigh--Bénard Convection (RBC) \citep{rayleigh1916, wen2022} describes the dynamics of a horizontal fluid layer heated from below and cooled from above, where buoyancy-driven instabilities generate convection currents with rising warm fluid and sinking cold fluid, governed by
\begin{equation}
    \frac{\partial b}{\partial t} - \kappa \Delta b = -u \cdot \nabla b,
\end{equation}
\begin{equation}
    \frac{\partial u}{\partial t} - \nu \Delta u + \nabla p - b\mathbf{e}_z = -u \cdot \nabla u,
\end{equation}
where $b$ denotes buoyancy, $u$ the velocity field, and $p$ the pressure. The thermal diffusivity $\kappa$ and viscosity $\nu$ are parameterized by the Rayleigh and Prandtl numbers as
\begin{equation}
    \kappa = (\text{Ra} \cdot \text{Pr})^{-1/2}, \quad
    \nu = (\text{Ra}/\text{Pr})^{-1/2}.
\end{equation}
The system exhibits strongly nonlinear and chaotic behavior, transitioning between laminar and turbulent regimes, making it a challenging benchmark for PDE emulation. The 2D state is represented by two scalar fields (buoyancy and pressure) and one vector field (velocity components), discretized on a $512 \times 128$ grid ($C=4$ channels). We gather data from The Well, selecting all the 200 timesteps ($\Delta t = 1$) available to evaluate the models along long rollouts. Each trajectory is associated with a distinct combination of Rayleigh and Prandtl numbers, providing a diverse range of dynamical regimes for training and evaluation.

\paragraph{Active Matter.} 
Active matter~\citep{maddu2024} describes the dynamics of $N$ elongated active
particles of length $\ell$ and thickness $b$ ($\ell/b \gg 1$) immersed in a
Stokes fluid of volume $V$. Individual particles convert chemical energy into
mechanical work, generating persistent active stresses that give rise to complex,
non-equilibrium dynamics. In the continuum limit, these systems are described by a
kinetic theory for the distribution function $\Psi(\mathbf{x},\mathbf{p},t)$
governed by the Smoluchowski equation,
\begin{gather}
    \frac{\partial \Psi}{\partial t} + \nabla_\mathbf{x} \cdot (\dot{\mathbf{x}}\Psi) + \nabla_\mathbf{p} \cdot (\dot{\mathbf{p}}\Psi) = 0,
\end{gather}
which ensures particle number conservation. The conformational fluxes
$\dot{\mathbf{x}}$ and $\dot{\mathbf{p}}$ arise from single-particle dynamics in a
background flow $\mathbf{u}(\mathbf{x},t)$ and, for dense suspensions, take the
form
\begin{gather}
    \dot{\mathbf{x}} = \mathbf{u} - d_T \nabla_\mathbf{x} \log \Psi, \\
    \dot{\mathbf{p}} = (\mathbf{I} - \mathbf{pp})\cdot(\nabla \mathbf{u} + 2\zeta \mathbf{D})\cdot \mathbf{p} - d_R \nabla_\mathbf{p} \log \Psi,
\end{gather}
where $d_T$ and $d_R$ are translational and rotational diffusion coefficients,
$\zeta$ controls alignment through steric interactions, and
$\mathbf{D} = \langle \mathbf{pp} \rangle$ is the second-moment tensor. Macroscopic
fields are obtained as moments of $\Psi$: the concentration $c=\langle 1 \rangle$,
polarity $\mathbf{n}=\langle \mathbf{p} \rangle / c$, and nematic order parameter
$\mathbf{Q} = \langle \mathbf{pp} \rangle / c$, with
$\langle f \rangle = \int_{|\mathbf{p}|=1} f \Psi \, d\mathbf{p}$. The particle
dynamics are coupled to the surrounding incompressible Stokes flow,
\begin{gather}
    -\Delta \mathbf{u} + \nabla P = \nabla \cdot \boldsymbol{\Sigma}, \qquad \nabla \cdot \mathbf{u} = 0,
\end{gather}
where the stress tensor
\begin{gather}
    \boldsymbol{\Sigma} = \alpha \mathbf{D} + \beta \mathbf{S} : \mathbf{E}
    - 2\zeta\beta (\mathbf{D} \cdot \mathbf{D} - \mathbf{S} : \mathbf{D})
\end{gather}
captures contributions from active dipoles ($\alpha$, the dipole strength), particle interactions ($\beta$, the particle density), and steric torques. Here $\mathbf{E} = (\nabla \mathbf{u} + \nabla \mathbf{u}^\top)/2$ is the rate-of-strain tensor and $\mathbf{S} = \langle \mathbf{pppp} \rangle$ the fourth-moment tensor. The dataset comprises 225 trajectories (5 random initializations for each of the 45 parameter sets) on a uniform $256\times256$ Cartesian grid with domain size $L_x = L_y = 10$ and periodic boundary conditions, stored at 81 snapshots separated by $\Delta t = 0.25$ seconds over a total window of $t \in [0, 20]$ s. The available fields are concentration, velocity components, orientation tensor, and strain-rate tensor. The parameter space covers active dipole strengths $\alpha \in \{-1,-2,-3,-4,-5\}$ and alignment coefficients $\zeta \in \{1,3,5,7,9,11,13,15,17\}$ at fixed $\beta=0.8$, spanning regimes from disordered isotropic motion to coherent nematic collective flows, and constituting a challenging benchmark for surrogate modelling of long-horizon non-equilibrium dynamics.

\clearpage
\section{Models}\label{si-sec:baselines}
This section details the model settings and hyperparameters used for training and inference. Unless stated otherwise, hyperparameters are shared across all PDE benchmarks; for WFM, the only system-specific choice is the mother wavelet (Table~\ref{si-tab:optimal_wavelets}). FNO and TFNO were retrained with the hyperparameters pre-tuned by The Well~\citep{ohana2024}. WNO and WDNO follow their original papers, with the network width enlarged to approach the parameter budget of WFM and the adaptations to autoregressive emulation described in the corresponding paragraphs below; for WNO, all fine-scale wavelet coefficients are retained rather than discarded. At inference time, generative models produce $\mathcal{M}=8$ trajectories ($\mathcal{M}=5$ for WDNO) from the same initial condition; at every autoregressive step, each trajectory draws fresh Gaussian noise and is integrated with the learned ODE for $N=50$ steps (flow-matching models) or sampled with deterministic DDIM for $N=4$ steps (WDNO).

\paragraph{How the baselines differ.} The generative baselines isolate the choice of representation. FM$_{\text{pixel}}$ uses the same flow-matching objective, sampler and U-Net as WFM-1 (12.7M parameters each) but generates in physical space. FM$_{\text{fourier}}$ generates the leading Fourier modes and discards the rest. WDNO denoises a whole space--time block of wavelet coefficients with a diffusion model. LatentFM$_{\div 64}$ generates in the latent space of a pre-trained autoencoder with a 64-fold compression. WFM-3 (30.6M) is compared with FM$_{\text{fourier}}$ (30.3M) at a matched parameter budget.

\subsection{Velocity-predicting U-Net}\label{sec:v_pred_arch}
\begin{figure}[ht!]
  \centering
  %\fbox{\rule[-.5cm]{0cm}{4cm} \rule[-.5cm]{4cm}{0cm}}
  \includegraphics[width=1.0\textwidth]{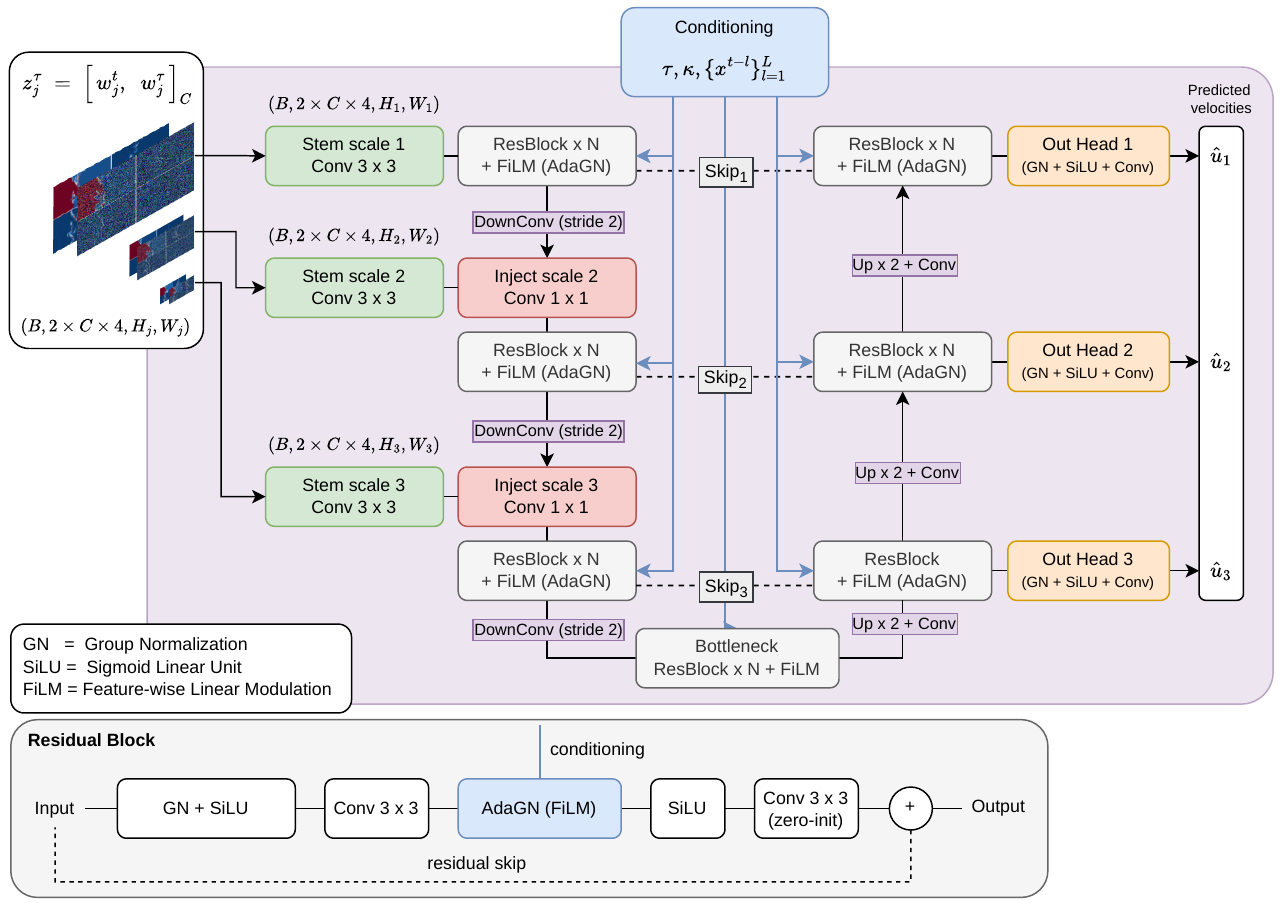}
  \caption{Illustration of the velocity-predicting U-Net $u_{\theta}$.}
  \label{fig:unet_diagram}
\end{figure}
The velocity field $u_\theta$ is parameterized by a U-Net (depicted in Figure~\ref{fig:unet_diagram}) whose depth and wavelet injection points are controlled independently by the number of wavelet scales ($n_\text{scales}$), and the number of encoder/decoder levels ($n_\text{levels}$), with $n_\text{levels} \geq n_\text{scales}$. This decoupling allows the same architecture to be used across all experimental conditions without structural changes.

\paragraph{Stems and injection.}
Each wavelet scale $j$ is processed by a dedicated \emph{stem}, a single $3\times3$ convolution that projects the channel-wise concatenation of the current-state coefficients $w_j^t$ and the noisy interpolant $w_j^{\tau}$ into the encoder feature space. The stem at scale $j=1$ initializes the encoder, while stems at scales $j > 1$ are fused into the encoder via a $1\times1$ convolution after the corresponding downsampling step, injecting multi-scale structure progressively into the representation.

\paragraph{Encoder, bottleneck, and decoder.}
Each encoder level applies three pre-activation residual blocks followed by a stride-2 convolution that halves the spatial resolution and doubles the channel width, up to a maximum of $8 \times C_\text{base}$. The bottleneck operates at the coarsest resolution with at least two residual blocks, providing a wide effective receptive field.
The decoder mirrors the encoder with bilinear upsampling followed by a $3\times3$ convolution, and restores resolution by concatenating skip connections from the corresponding encoder level. A lightweight output head (GroupNorm--SiLU--Conv) is applied at each decoder level that corresponds to a wavelet scale, producing one velocity prediction per scale ($\hat{u}_j$).

\paragraph{Conditioning.}
All residual blocks use AdaGN~\citep{dhariwal2021}: the conditioning vector $\mathbf{c} \in \mathbb{R}^{d}$, obtained by summing a sinusoidal flow-time embedding, constant parameters of the PDE ($\kappa$), and the spatiotemporal context embedding of the past $L=3$ states, modulates the feature activations via learned scale and shift parameters. Throughout the experiments we set $d=256$.

\paragraph{Special cases.}
Setting $n_\text{scales} = n_\text{levels}$ recovers a standard multi-scale wavelet U-Net in which every encoder level receives the wavelet representation at the appropriate scale. Setting $n_\text{scales} = 1$ disables all injections beyond the input stem, reducing the model to a plain U-Net operating on the finest wavelet scale only ($j=1)$. This is the configuration used for WFM 1-scale variants. Instead, FM$_{pixel}$ uses the same architecture with wavelet injection disabled, and the stem operating directly on the full pixel grid.

\paragraph{Cross-scale coupling.}\label{si-sec:cross_scale}
Because every head $\hat{u}_j$ is computed by the shared encoder--decoder, the $J$ flows are not independent. We quantify this coupling for WFM 3-scale (haar) on TRL. If the flows were independent, each head would depend only on its own scale, i.e.\ $\partial \hat{u}_j / \partial w^{\tau}_{j'} = 0$ for $j \neq j'$. We therefore perturb the interpolant $w^{\tau}$ at a single scale $j'$ and measure the relative change of the velocity predicted by every head $j$:
\begin{equation}
  S_{jj'} = \frac{1}{P}\sum_{p=1}^{P}
  \frac{\bigl\lVert \hat{u}_j\bigl(w^{\tau} + \delta\, v^{(p)}_{j'}\bigr) - \hat{u}_j\bigl(w^{\tau}\bigr) \bigr\rVert_{\mathrm{RMS}}}
       {\delta\, \bigl\lVert \hat{u}_j\bigl(w^{\tau}\bigr) \bigr\rVert_{\mathrm{RMS}}}.
  \label{eq:sensitivity}
\end{equation}
Here $v^{(1)}_{j'}, \dots, v^{(P)}_{j'}$ are $P=16$ random perturbation directions, each drawn from $\mathcal{N}(0, I)$ on the coefficients of scale $j'$, set to zero on all other scales, and rescaled to the RMS of $w^{\tau}_{j'}$. The scalar $\delta$ sets the perturbation size relative to that RMS; we use $\delta = 10^{-2}$. $S_{jj'}$ is thus the relative change of head $j$ per unit relative perturbation of scale $j'$. We average it over $8$ time windows of a test trajectory and five interpolation times $\tau \in \{0.1, 0.3, 0.5, 0.7, 0.9\}$, and summarize the resulting matrix by its off-diagonal mass $\rho = \sum_{j \neq j'} S_{jj'} / \sum_{j,j'} S_{jj'}$, which is exactly zero for independent flows.

We obtain $\rho = 15.1\%$ (Table~\ref{tab:sensitivity}), and between $14.5\%$ and $15.3\%$ at each individual $\tau$. At $\tau = 0.5$, repeating the measurement with $\delta = 10^{-3}$ and $\delta = 10^{-1}$ changes $\rho$ by at most $0.3$ percentage points, so the result does not depend on the perturbation size. Evaluating the network twice on identical inputs gives a difference of zero to machine precision, so the off-diagonal entries are not numerical noise. The coupling runs in both directions: coarse-to-fine entries (above the diagonal) average $0.271$, and fine-to-coarse entries (below the diagonal) $0.138$. This measures the local sensitivity of the learned velocity field; it does not show that the coupling reproduces the correct joint statistics across scales.

\begin{table}[ht]
  \centering
  \caption{Cross-scale sensitivity $S_{jj'}$ (Equation~\ref{eq:sensitivity}) of WFM 3-scale (haar) on TRL, with $\delta = 10^{-2}$, averaged over test windows and $\tau$. Rows: affected head $\hat{u}_j$; columns: perturbed scale $w^{\tau}_{j'}$ ($j=1$ finest). Independent per-scale flows would give zero off-diagonal entries; the measured off-diagonal mass is $\rho = 15.1\%$.}
  \label{tab:sensitivity}
  \begin{tabular}{lccc}
    \toprule
    & $w^{\tau}_1$ (fine) & $w^{\tau}_2$ (mid) & $w^{\tau}_3$ (coarse) \\
    \midrule
    $\hat{u}_1$ (fine)   & 2.292 & 0.225 & 0.277 \\
    $\hat{u}_2$ (mid)    & 0.135 & 2.366 & 0.311 \\
    $\hat{u}_3$ (coarse) & 0.095 & 0.183 & 2.243 \\
    \bottomrule
  \end{tabular}
\end{table}

%\clearpage
\subsection{Wavelet Flow Matching}
To ensure a fair comparison across scale configurations, we normalize model capacity. The 3-scale WFM employs a shared U-Net backbone with three output heads, making it an inherently harder multi-target task compared to single-scale prediction. In U-Net architectures, capacity is primarily driven by the convolutions in residual blocks ~\citep{he2016}, which scale as $\mathcal{O}(\text{init\_dim}^2)$, where $\text{init\_dim}$ denotes the number of feature channels in the first layer, assuming input and output channels grow proportionally to $\text{init\_dim}$. We therefore equalize capacity on a per-head basis. In particular, for $\text{init\_dim}=64$, the per-head capacity is proportional to $64^2/3 \approx 1365$, which corresponds to $\text{init\_dim}  = 64 / \sqrt{3} \approx 37$ for single-head models (e.g., WFM 1-scale). We round this value to $\text{init\_dim} = 40$, the nearest multiple of $8$ above $37$, to satisfy divisibility by $8$ of GroupNormalization layers. WFM 1-scale and FM$_{pixel}$ therefore use $\text{init\_dim}=40$, while FM$_{fourier}$ uses $\text{init\_dim}=64$ to match WFM 3-scale. 

%All single-head baselines are therefore retrained with $\text{init\_dim}=40$.

\begin{table}[ht!]
\centering
\caption{Hyperparameters for the WFM configurations.}
\label{tab:config_wfm}
\begin{tabular}{lcc}
\toprule
 & \multicolumn{1}{c}{WFM 1-scale} & \multicolumn{1}{c}{WFM 3-scale} \\
\midrule
Architecture         & U-Net                    & U-Net                    \\
Initial dimension    & 40                       & 64                       \\
Residual blocks      & 3                        & 3                        \\
Embedding dimension  & 256                      & 256                      \\
Parameters           & $12.7 \times 10^6$       & $30.6 \times 10^6$       \\
\midrule
Wavelet scales          & 1              & 3              \\
Mother wavelet          & PDE-specific   & PDE-specific   \\
Padding mode        & periodic & periodic \\
Scales weighting scheme & 1.0 for all scales        & 1.0 for all scales        \\
\midrule
Optimizer          & AdamW                  & AdamW                  \\
Learning rate      & $3\times10^{-4}$       & $3\times10^{-4}$       \\
Weight decay       & $1\times10^{-2}$       & $1\times10^{-2}$       \\
Betas              & $(0.9,\,0.999)$        & $(0.9,\,0.999)$        \\
Scheduler          & CosineAnnealingLR      & CosineAnnealingLR      \\
Eta min            & $1\times10^{-7}$       & $1\times10^{-7}$       \\
Epochs             & 200                    & 200                    \\
Warm-up epochs     & 20                     & 20                     \\
Batch size         & 32                     & 32                     \\
Gradient clipping  & 1.0                    & 1.0                    \\
Precision          & \texttt{bf16-mixed}    & \texttt{bf16-mixed}    \\
Matmul precision   & \texttt{high}          & \texttt{high}          \\
Strategy           & DDP                    & DDP                    \\
Framework          & PyTorch Lightning      & PyTorch Lightning      \\
GPUs               & $4\times$ A100 (40GB)  & $4\times$ A100 (40GB)  \\
\bottomrule
\end{tabular}
\end{table}

\clearpage
\subsection{Flow Matching in Pixel Space}

\begin{table}[ht]
\centering
\caption{Hyperparameters for the FM$_{pixel}$ configuration.}
\label{tab:config_fm_pixel}
\begin{tabular}{lc}
\toprule
 & \multicolumn{1}{c}{FM$_{pixel}$} \\
\midrule
Architecture         & U-Net                \\
Initial dimension    & 40                   \\
Residual blocks      & 3                    \\
Embedding dimension  & 256                  \\
Parameters           & $12.7 \times 10^6$   \\
\midrule
Optimizer          & AdamW                  \\
Learning rate      & $6\times10^{-4}$       \\
Weight decay       & $1\times10^{-2}$       \\
Betas              & $(0.9,\,0.999)$        \\
Scheduler          & CosineAnnealingLR      \\
Eta min            & $1\times10^{-7}$       \\
Epochs             & 200                    \\
Warm-up epochs     & 20                     \\
Batch size         & 16                     \\
Gradient clipping  & 1.0                    \\
Precision          & \texttt{bf16-mixed}    \\
Matmul precision   & \texttt{high}           \\
Strategy           & DDP                    \\
Framework          & PyTorch Lightning      \\
GPUs               & $4\times$ A100 (40GB)  \\
\bottomrule
\end{tabular}
\end{table}

\subsection{Flow Matching in Fourier Space}
\begin{table}[ht]
\centering
\caption{Hyperparameters for the FM$_{fourier}$ configuration.}
\label{tab:config_fm_fourier}
\begin{tabular}{lc}
\toprule
 & \multicolumn{1}{c}{FM$_{fourier}$} \\
\midrule
Architecture         & U-Net                \\
Initial dimension    & 64                   \\
Residual blocks      & 3                    \\
Embedding dimension  & 256                  \\
Parameters           & $30.3 \times 10^6$   \\
\midrule
Fourier Modes (H, W) & (32, 32)             \\
\midrule
Optimizer          & AdamW                  \\
Learning rate      & $1\times10^{-4}$       \\
Weight decay       & $1\times10^{-2}$       \\
Betas              & $(0.9,\,0.999)$        \\
Scheduler          & CosineAnnealingLR      \\
Eta min            & $1\times10^{-7}$       \\
Epochs             & 200                    \\
Warm-up epochs     & 20                     \\
Batch size         & 32                     \\
Gradient clipping  & 1.0                    \\
Precision          & \texttt{bf16-mixed}    \\
Matmul precision   & \texttt{high}           \\
Strategy           & DDP                    \\
Framework          & PyTorch Lightning      \\
GPUs               & $4\times$ A100 (40GB)  \\
\bottomrule
\end{tabular}
\end{table}

\newpage
\subsection{Wavelet Neural Operator}
Our implementation keeps the core WNO operator structure~\cite{tripura2023wno},
adapting the wrapper to the WFM benchmark: we flatten history and field channels
as model input, append available physical scalar conditions through the benchmark
data interface, and evaluate using identical WFM normalization and rollout
pipeline. For Rayleigh--Bénard we apply a logarithmic transform to the physical
parameters; for TRL and active matter we use the default scalar representation.
Hyperparameters are reported in Table~\ref{tab:config_wno}.

\begin{table}[ht!]
\centering
\caption{Hyperparameters for the WNO configuration.}
\label{tab:config_wno}
\begin{tabular}{lc}
\toprule
 & WNO \\
\midrule
Hidden width        & 32       \\
Layers              & 4        \\
Wavelet scales      & 3        \\
Mother wavelet      & $\mathrm{db}2$ \\
Padding mode        & periodic \\
Input steps         & 4        \\
Output steps        & 1        \\
\midrule
Optimizer               & AdamW             \\
Learning rate           & $1\times10^{-3}$  \\
Weight decay            & $1\times10^{-6}$  \\
Betas                   & $(0.9,\,0.999)$   \\
Scheduler               & CosineAnnealingLR \\
Eta min                 & $1\times10^{-6}$  \\
Epochs                  & 200               \\
Early-stopping patience & 20                \\
Batch size              & 16                \\
Framework               & PyTorch Lightning  \\
GPUs                    & $1\times$ A100 (40GB)  \\
\bottomrule
\end{tabular}
\end{table}

\clearpage
\subsection{Wavelet Diffusion Neural Operator}
Our WDNO baseline~\cite{hu2024wdno} is a diffusion model that operates in wavelet space. We adapt it to the $4\to1$ setting used here as follows. The input to the model is the full spatiotemporal block $[t_1, t_2, t_3, t_4, t_5]$, transformed via a 3D DWT over time and space; a 3D U-Net is then trained to denoise in this wavelet state space. For conditioning, the observed steps $[t_1, t_2, t_3, t_4]$ are provided as fixed channels, with known components re-imposed at each sampling step. We use the $\mathrm{db2}$ mother wavelet with periodic padding, $u$-prediction, deterministic DDIM sampling, and EMA weights at inference. Full hyperparameters are reported
in Table~\ref{tab:config_wdno}.

\begin{table}[ht!]
\centering
\caption{Hyperparameters for the WDNO configuration.}
\label{tab:config_wdno}
\begin{tabular}{lc}
\toprule
 & WDNO \\
\midrule
Architecture        & U-Net \\
Initial dimension   & 64      \\
Residual blocks          & 1       \\
U-Net resolution levels  & 3       \\
Conditioning dimension   & 128     \\
Input steps     & 4   \\
Output steps     & 1   \\
\midrule
Mother wavelet      & $\mathrm{db}2$  \\
Padding mode        & periodic        \\
Wavelet scales      & 1               \\
\midrule
Diffusion steps     & 4                   \\
Sampling steps      & 4                   \\
Sampler             & DDIM ($\eta=0$)     \\
Beta schedule       & Cosine              \\
$x_0$ clipping      & $[-1,\,1]$          \\
EMA decay           & 0.995               \\
\midrule
Optimizer               & AdamW             \\
Learning rate           & $1\times10^{-4}$  \\
Weight decay            & $1\times10^{-6}$  \\
Betas                   & $(0.9,\,0.999)$   \\
Scheduler               & CosineAnnealingLR \\
Eta min                 & $1\times10^{-6}$  \\
Epochs                  & 200               \\
Early-stopping patience & 20                \\
Batch size              & 4                 \\
Precision               & \texttt{fp32}     \\
Framework               & PyTorch Lightning  \\
GPUs                    & $1\times$ A100 (40GB)  \\
\bottomrule
\end{tabular}
\end{table}

\clearpage
\subsection{Fourier-based Neural Operators}

\begin{table}[ht]
\centering
\caption{Hyperparameters for the FNO and TFNO configurations.}
\label{tab:config_fno}
\begin{tabular}{lcc}
\toprule
 & \multicolumn{1}{c}{FNO} & \multicolumn{1}{c}{TFNO} \\
\midrule
Architecture         & FNO                    & TFNO                   \\
Hidden dimension     & 128                    & 128                    \\
Fourier blocks       & 4                      & 4                      \\
Parameters           & $19\times10^{6}$       & $7.9\times10^{6}$       \\
\midrule
Fourier Modes (H, W) & (16, 16)               & (16, 16)               \\
Tucker rank          & -                      & 0.42                    \\
\midrule
Learning rate              & $5\times10^{-3}$       & $5\times10^{-3}$       \\
Epochs                     & 500                    & 500                    \\
Batch size                 & 16                     & 16                     \\
Early stopping patience    & 20                     & 20                     \\
Early stopping min delta   & $1\times10^{-6}$       & $1\times10^{-6}$       \\
Precision                  & \texttt{32-true}       & \texttt{32-true}       \\
Matmul precision           & \texttt{high}          & \texttt{high}          \\
Strategy                   & DDP                    & DDP                    \\
Framework                  & PyTorch Lightning      & PyTorch Lightning      \\
GPUs                       & $2\times$ A100 (40GB)  & $2\times$ A100 (40GB)  \\
\bottomrule
\end{tabular}
\end{table}

\subsection{Latent Flow Matching}\label{si-sec:latentfm}
%We follow \citet{rozet2025} and use their pre-trained DC-AE (f32c64) for Rayleigh--B\'enard, which compresses each $512\times128\times4$ state into a $16\times4\times64$ latent ($\div 64$). Flow matching is trained in this latent space with the same OT-CFM objective as WFM; the hyperparameters are listed in Table~\ref{si-tab:latentfm}. Only the flow model is trained ($8.4\times10^6$ parameters); the autoencoder ($313.1\times10^6$ parameters) is kept frozen, and its encoder and decoder are called once per autoregressive step. The autoencoder was pre-trained by \citet{rozet2025} (about one day on 8 H100 GPUs); WFM requires no pre-training stage and trains in about 6 hours on 4 A100 GPUs. LatentFM is evaluated on Rayleigh--B\'enard only, the system for which a pre-trained autoencoder is available.

LatentFM$_{\div 64}$ follows \citet{rozet2025} and is evaluated on Rayleigh--B\'enard only, the system for which their pre-trained autoencoder is available. The autoencoder (DC-AE f32c64, $313.1\times10^6$ parameters) compresses each $512\times128\times4$ state into a $16\times4\times64$ latent, a 64-fold reduction, and is kept frozen; its encoder and decoder are called once per autoregressive step. Only the latent flow model is trained ($8.4\times10^6$ parameters), with the same OT-CFM objective as WFM and the hyperparameters in Table~\ref{si-tab:latentfm}. In terms of cost, the autoencoder required about one day of pre-training on 8 H100 GPUs~\citep{rozet2025}, whereas WFM needs no pre-training stage and trains in about 6 hours on 4 A100 GPUs.

\begin{table}[h]
\centering
\caption{LatentFM$_{\div 64}$ hyperparameters.}
\label{si-tab:latentfm}
\begin{tabular}{ll}
\toprule
Autoencoder & DC-AE f32c64, pre-trained~\citep{rozet2025}, frozen \\
Latent shape & $16\times4\times64$ \\
U-Net & base channels 64, 2 levels, 3 res.\ blocks, cond.\ dim 256 \\
Parameters & $8.4 \times 10^6$ trainable (flow model); $313.1 \times 10^6$ frozen (autoencoder) \\
Optimizer & AdamW, lr $10^{-4}$, weight decay $10^{-2}$ \\
Schedule & 20 warm-up epochs, cosine decay to $10^{-7}$ \\
Epochs / batch size & 200 / 32 \\
Precision & bf16, DDP \\
Sampler & Euler, $N=50$ steps, $\mathcal{M}=8$ trajectories \\
\bottomrule
\end{tabular}
\end{table}

\clearpage
\section{Metrics}\label{si-sec:metrics}
Let $u \in \mathbb{R}^{H \times W}$ denote the ground-truth field at a given rollout step, and $\langle \cdot \rangle$ the mean over the spatial dimensions. A generative model returns the predictions $v_1, \dots, v_{\mathcal{M}}$ of its $\mathcal{M}$ sampled trajectories at that step ($\mathcal{M}=8$, or $\mathcal{M}=5$ for WDNO). For simplicity of notation, we write $v \in \mathbb{R}^{H \times W}$ for the prediction: for deterministic models it is the model output, and for generative models it is the sample mean $\frac{1}{\mathcal{M}}\sum_{m} v_m$. CRPS is computed from the individual samples $v_m$, and for spectral coherence the power and cross-spectra are computed for each $v_m$ and then averaged over the samples.

%Let $u \in \mathbb{R}^{H \times W}$ and $v \in \mathbb{R}^{H \times W}$ denote the ground-truth and predicted spatial fields at a given rollout step. For generative models, $v_1, \dots, v_{\mathcal{M}}$ denote the predictions of the $\mathcal{M}$ sampled trajectories at that step ($\mathcal{M}=8$, or $\mathcal{M}=5$ for WDNO), and $v$ is their mean. Let $\langle \cdot \rangle$ denote the mean over the spatial dimensions.

\paragraph{Variance-normalized Root Mean Squared Error (VRMSE).}
The RMSE of the prediction $v$, normalized by the spatial standard deviation of the reference field,

\begin{equation}
    \text{VRMSE}(u, v) = \sqrt{\frac{\bigl\langle (u - v)^2 \big\rangle}{\bigl\langle (u-\langle u \rangle)^2\big\rangle+\epsilon}}
\end{equation}
where $\epsilon = 10^{-6}$ ensures numerical stability. A prediction equal to the spatial mean of $u$ yields $\mathrm{VRMSE} \approx 1$, providing a meaningful baseline~\cite{ohana2024, rozet2025}. Unlike the normalised RMSE (NRMSE), which divides by $\langle u^2\rangle$, VRMSE avoids down-weighting errors in non-negative fields (e.g.\ pressure and density)~\cite{rozet2025}.

\paragraph{Fair Continuous Ranked Probability Score (CRPS).} 
The CRPS generalizes the mean absolute error (MAE) to probabilistic forecasts by scoring an entire predictive distribution against an observation \cite{gneiting2007}. For $\mathcal{M}$ samples $\{v_m\}$, the unbiased ("fair") estimator of Ferro~\cite{ferro2014} is
\begin{equation}
  {\mathrm{CRPS}}_{\mathrm{fair}}(u,\{v_m\})
  \;=\;
  \frac{1}{\mathcal{M}}\sum_{m=1}^{\mathcal{M}}\langle|v_m - u|\rangle
  \;-\;
  \frac{1}{\mathcal{M}(\mathcal{M}-1)}\sum_{1\le m < m'\le \mathcal{M}}\langle|v_m - v_{m'}|\rangle,
  \label{eq:crps}
\end{equation}
The first term penalizes deviations of individual samples from the truth; the second rewards the spread between samples. Equation~\eqref{eq:crps} is an unbiased estimator of the CRPS of the distribution from which the samples are drawn~\cite{ferro2014}; the unadjusted $1/\mathcal{M}^2$ weight favours under-dispersed samples. Because the CRPS is a strictly proper scoring rule, its expected value is minimized when the samples are drawn from the true conditional distribution of $u$~\cite{gneiting2007}. For turbulent flows, CRPS measures whether the model correctly represents the range of possible realizations rather than just their mean, making it essential when trajectories diverge chaotically after a finite predictability horizon.

\paragraph{Spectral Coherence RMSE.}
Let $P_u(k)$ and $P_v(k)$ denote the isotropic power spectral densities
(PSDs) of fields $u$ and $v$ respectively, at radial wavenumber $k$
(cycles\,pixel$^{-1}$), obtained by azimuthal averaging of the 2-D DFT
magnitude squared.  Let $C_{uv}(k) = |\hat{u}(k)\,\overline{\hat{v}(k)}|$
be the corresponding cross-spectrum magnitude.  The spectral coherence at
wavenumber $k$ is
\begin{equation}
  \gamma(k)
  =
  \frac{C_{uv}(k) + \varepsilon}
       {\sqrt{P_u(k)\,P_v(k)} + \varepsilon},
  \label{eq:coherence}
\end{equation}
where $\varepsilon = 10^{-6}$.  By the Cauchy--Schwarz inequality, $\gamma(k)\in[0,1]$. We distribute the resulting wavenumbers evenly in log-space and divide into three frequency bands (\emph{Low}, \emph{Mid}, \emph{High}), then for each band:
\begin{equation}
  \mathrm{RMSE}_{c}(\mathcal{B})
  =
  \sqrt{
    \frac{1}{|\mathcal{B}|}
    \sum_{k\in\mathcal{B}}
    \bigl(1 - \gamma(k)\bigr)^{2}
  }.
  \label{eq:rmsec}
\end{equation}
$\mathrm{RMSE}_c\!\approx\!0$ indicates that $v$ reproduces the
correct phase and amplitude structure of $u$ at each resolved scale
within the band; $\mathrm{RMSE}_c\!=\!1$ (the maximum, by
Cauchy--Schwarz) indicates the two fields are spectrally uncorrelated
across the band.  For turbulent flows energy cascades from large to small scales, so coherence typically degrades first at high wavenumbers; $\mathrm{RMSE}_c$ therefore quantifies whether the model respects the inertial-range cascade structure at each resolved scale, complementing the point-wise VRMSE which loses sensitivity once trajectories have diverged chaotically. We prefer Spectral Coherence RMSE over Power Spectrum RMSE, which is defined as in Equation~\eqref{eq:rmsec} but replacing $\gamma(k)$ with the ratio $P_v(k) / P_u(k)$. This choice is motivated by the fact that Power Spectrum RMSE is unbounded and numerically unstable when $P_u(k) \approx 0$, which is the case of near-zero-mean fields in some PDEs considered in this study.

\clearpage
\section{Mother Wavelets}\label{si-sec:wavelets}
All four wavelets used in this study belong to the Daubechies family \citep{daubechies1988}, which constructs compactly supported orthonormal bases of $L^{2}(\mathbb{R})$. A family $\{\psi_{j,n}\}_{(j,n)\in\mathbb{Z}^{2}}$ is \emph{orthonormal} when
\begin{equation}
    \langle \psi_{j,n},\, \psi_{j',n'} \rangle = \delta_{jj'}\,\delta_{nn'},
    \label{eq:orthonormal}
\end{equation}
with $\delta_{ij}$ the Kronecker delta ($1$ if $i=j$, $0$ otherwise). This means that the basis functions are both mutually orthogonal and of unit norm. This is strictly stronger than mere orthogonality, where cross inner products vanish but norms need not equal one. Because the basis is orthonormal, the $L^{2}$ energy of any signal is exactly preserved in its coefficient sequence (Parseval's identity), so energy-based metrics such as RMSE carry the same meaning in both the signal and the wavelet domain without any additional rescaling.
 
\paragraph{Vanishing moments and polynomial cancellation.}
The Daubechies $\mathrm{db}p$ wavelet is the unique compactly supported,
orthonormal wavelet that maximises the number of vanishing moments $p$ for a
support of length $2p - 1$ (equivalently, a filter of length $2p$).
Having $p$ vanishing moments means
\begin{equation}
    \int_{-\infty}^{\infty} t^{k}\,\psi(t)\,\mathrm{d}t = 0,
    \qquad k = 0, \ldots, p-1,
    \label{eq:vanishing}
\end{equation}
so the wavelet transform exactly annihilates polynomials of degree less than
$p$: detail coefficients are identically zero wherever the signal is locally
well approximated by such polynomials.
Increasing $p$ therefore yields sparser coefficient representations for smooth
signals, at the cost of a wider spatial support and a slightly larger
convolution per decomposition level.
 
The Haar wavelet (equivalent to $\mathrm{db}1$) is the simplest member of this family: its
scaling function is piecewise constant on $[0,1)$, giving a support of length
one and a single vanishing moment. It is the only Daubechies wavelet that is symmetric and discontinuous, making it computationally cheap but poorly suited to capturing smooth transitions. Instead, $\mathrm{db}2$, $\mathrm{db}4$, and $\mathrm{db}6$ progressively widen the
support and increase regularity, suppressing polynomial trends of degree up to
1, 3, and 5, respectively. Table~\ref{tab:wavelets} summarises the key properties of the four mother wavelets considered in this work.
 
%\paragraph{Comparison with other orthonormal families.}
%Two closely related families are worth distinguishing. \emph{Symlets} ($\mathrm{sym}p$) are near-symmetric modifications of $\mathrm{db}p$: they share identical support length and vanishing-moment order but minimise the asymmetry of the phase response, reducing edge artefacts at the cost of a slightly relaxed exactness in symmetry \citep{daubechies1992ten}. \emph{Coiflets} additionally impose $p$ vanishing moments on the \emph{scaling function} $\phi$ itself (not only on $\psi$), which improves the accuracy of approximation coefficients at a given scale, but requires a filter roughly three times longer than the corresponding Daubechies wavelet. All three families yield orthonormal bases; their main differences lie in support width, phase linearity, and where the vanishing-moment constraints are placed.

%For reference: symlets ($\mathrm{sym}p$) match $\mathrm{db}p$ in support and vanishing moments but achieve near-symmetry; coiflets additionally enforce vanishing moments on $\phi$ at the cost of a filter $\approx\!3\times$ longer.

\begin{table}[h]
\centering
\caption{Key properties of the mother wavelets used in this study. All four belong to the Daubechies ($\mathrm{db}p$) family \citep{daubechies1988, daubechies1992} and form orthonormal bases of $L^{2}(\mathbb{R})$. Filter length equals $2p$ and support length equals $2p-1$ for $\mathrm{db}p$.}
\label{tab:wavelets}
\smallskip
\begin{tabular}{lccccc}
\toprule
\textbf{Wavelet} &
\makecell{\textbf{Filter} \\ \textbf{length}} &
\textbf{Support} &
\makecell{\textbf{Vanishing} \\ \textbf{moments}} &
\makecell{\textbf{Polynomials} \\ \textbf{annihilated}} &
\textbf{Symmetry} \\
\midrule
$\mathrm{db}1$ (haar) & $2$  & $[0,\,1]$  & $1$ & constants $(\deg \le 0)$  & symmetric  \\
$\mathrm{db}2$        & $4$  & $[0,\,3]$  & $2$ & linear    $(\deg \le 1)$  & asymmetric \\
$\mathrm{db}4$        & $8$  & $[0,\,7]$  & $4$ & cubic     $(\deg \le 3)$  & asymmetric \\
$\mathrm{db}6$        & $12$ & $[0,\,11]$ & $6$ & quintic   $(\deg \le 5)$  & asymmetric \\
\bottomrule
\end{tabular}
\end{table}

\paragraph{Discrete Wavelet Transform Implementation.}
Our DWT is implemented in PyTorch using the \texttt{PyWavelets} (\texttt{pywt})~\citep{lee2019} backend, using the \texttt{periodic} padding mode. The transform operates on the spatial dimensions $(H, W)$ only, leaving the batch, channel, and temporal axes untouched. To enable batched processing, input tensors of shape $(B, C, T, H, W)$ are transiently reshaped to $(B \cdot C \cdot T,\, H, W)$ before each single-level DWT call and restored afterwards.

\clearpage
\section{Additional Experimental Results}\label{si-sec:results}
\begin{figure}[ht!]
  \centering
  %\fbox{\rule[-.5cm]{0cm}{4cm} \rule[-.5cm]{4cm}{0cm}}
  \includegraphics[width=1.0\textwidth]{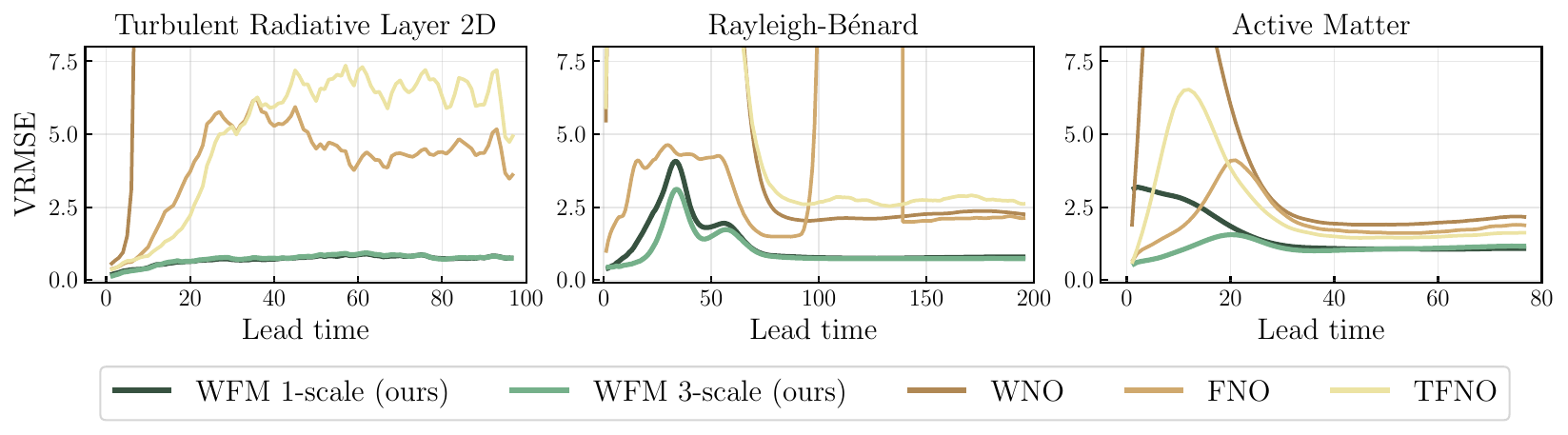}
  \caption{VRMSE across rollout steps for WFM and deterministic baselines per PDE system.}
  \label{si-fig:vrmse}
\end{figure}

\begin{figure}[h]
  \centering
  %\fbox{\rule[-.5cm]{0cm}{4cm} \rule[-.5cm]{4cm}{0cm}}
  \includegraphics[width=1.0\textwidth]{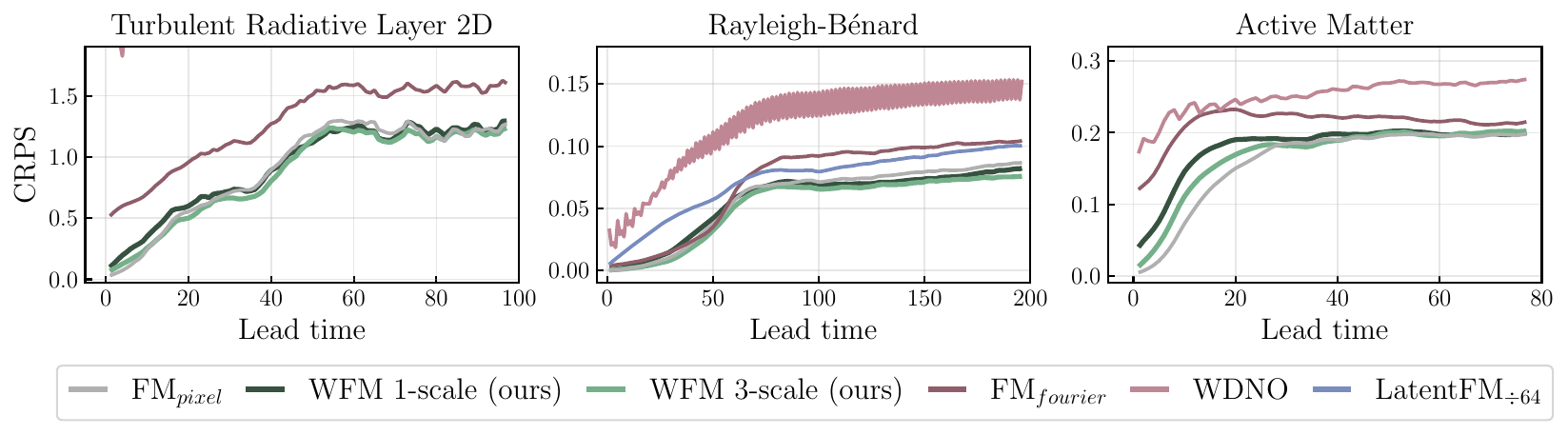}
  \caption{CRPS across rollout steps for WFM and generative baselines per PDE system.}
  \label{si-fig:crps}
\end{figure}

\begin{figure}[h]
  \centering
  %\fbox{\rule[-.5cm]{0cm}{4cm} \rule[-.5cm]{4cm}{0cm}}
  \includegraphics[width=1.0\textwidth]{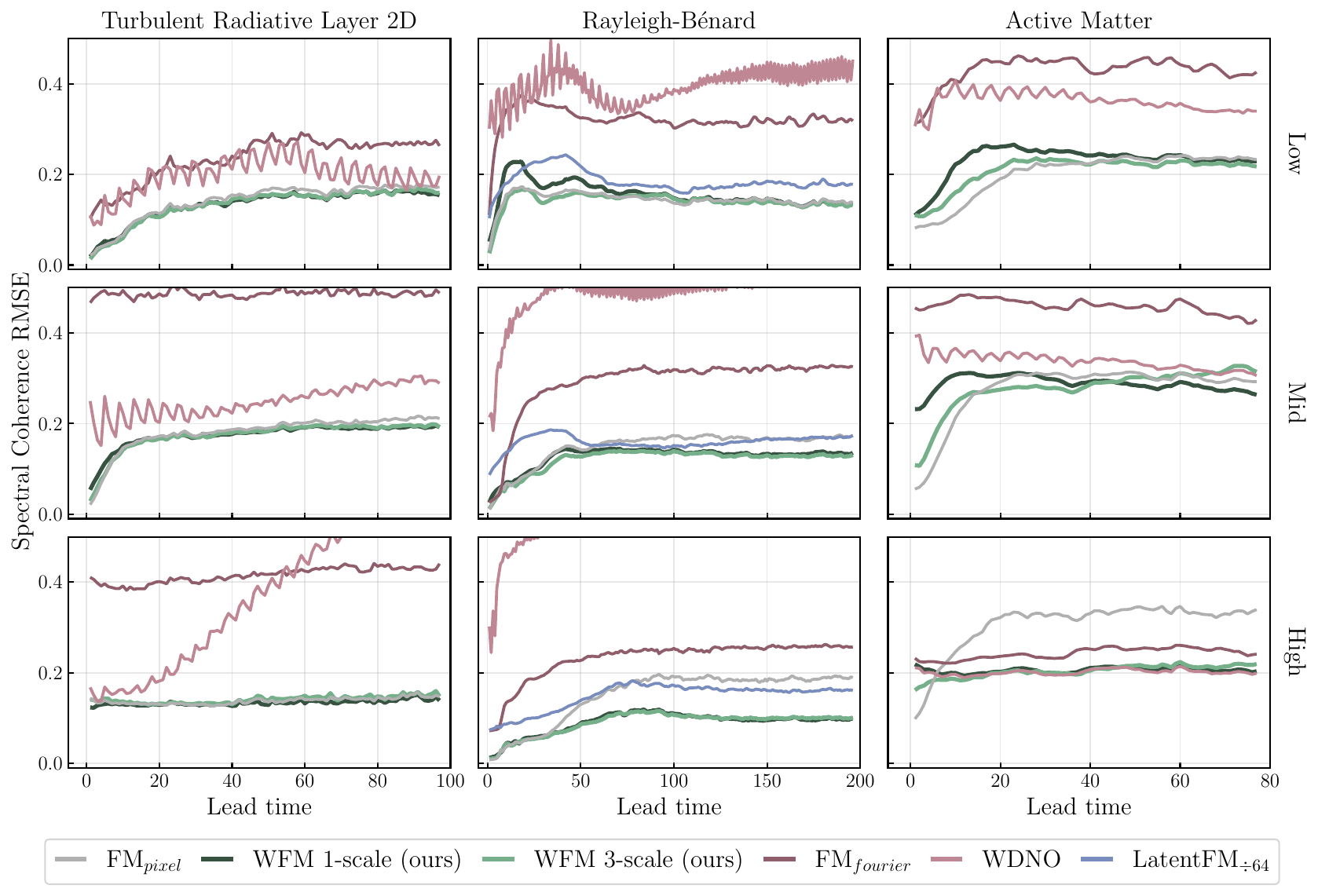}
  \caption{Spectral coherence RMSE across frequency bands and rollout steps for WFM and generative baselines per PDE system.}
  \label{si-fig:spec_coherence}
\end{figure}

This section provides additional material and analyses complementing the main text. Figures~\ref{si-fig:vrmse}, \ref{si-fig:crps}, and \ref{si-fig:spec_coherence} show WFM variants based on the wavelet selection reported in Section~\ref{sec:wfm_eval} and summarized in Table~\ref{si-tab:optimal_wavelets}. Bold and underlined entries denote the best and second-best results, respectively. All metrics reported in this section are lower-is-better.

\begin{table}[ht]
\centering
\caption{Optimal mother wavelet per WFM variant and PDE system, selected by lowest spectral coherence RMSE averaged across \textit{Low}, \textit{Mid}, and \textit{High} frequency bands and rollout horizons (Table~\ref{tab:avg_sc_rmse_gen_wavelets}). In parentheses: the selected wavelet's score, mean\,$\pm$\,std over three inference seeds (42, 84, 168); the selection is made on the seed mean.}
\label{si-tab:optimal_wavelets}
\begin{tabular}{lccc}
\toprule
 & Turbulent Radiative Layer 2D & Rayleigh--B\'{e}nard & Active Matter \\
\midrule
WFM 1-scale & $\mathrm{haar}$ (0.147{\scriptsize$\,\pm\,$$<$0.001}) & $\mathrm{db}4$ (0.121{\scriptsize$\,\pm\,$0.002}) & $\mathrm{haar}$ (0.241{\scriptsize$\,\pm\,$0.001}) \\
WFM 3-scale & $\mathrm{haar}$ (0.149{\scriptsize$\,\pm\,$$<$0.001}) & $\mathrm{db}4$ (0.109{\scriptsize$\,\pm\,$0.001}) & $\mathrm{db}2$ (0.220{\scriptsize$\,\pm\,$$<$0.001}) \\
\bottomrule
\end{tabular}
\end{table}

%\newpage
\paragraph{VRMSE across autoregressive rollout steps.} \quad
\begin{table}[ht]
\centering
\caption{VRMSE across rollout horizons for all models and PDE systems. Deterministic baselines (WNO, FNO, TFNO) are single runs. LatentFM$_{\div 64}$ is evaluated on Rayleigh--B\'enard only. Mean over three inference seeds (42, 84, 168); the standard deviation across seeds is below 5\% of the mean for all entries except those marked $^{*}$. Here, inf indicates that the rollout diverged, while -- indicates that the model was not available for the corresponding system.}
\label{si-tab:appendix_vrmse}
\resizebox{\linewidth}{!}{%
\begin{tabular}{l ccc ccc ccc}
\toprule
 & \multicolumn{3}{c}{Turbulent Radiative Layer 2D} & \multicolumn{3}{c}{Rayleigh--B\'{e}nard} & \multicolumn{3}{c}{Active Matter} \\
\cmidrule(lr){2-4} \cmidrule(lr){5-7} \cmidrule(lr){8-10}
Rollout & 01:32 & 33:64 & 65:97 & 01:32 & 33:64 & 65:196 & 01:16 & 17:32 & 33:77 \\
\midrule
WFM-1 haar & \underline{0.544} & 0.791 & 0.789 & 1.782 & 2.020 & 0.788 & 2.847 & 1.541 & 1.079 \\
WFM-1 db2 & 0.550 & 0.750 & 0.850 & 1.333 & 1.844$^{*}$ & 0.791 & 5.652 & 2.930 & 1.622 \\
WFM-1 db4 & 0.550 & \underline{0.730} & 0.811 & 1.689 & 2.271 & 0.802 & 14.227 & 5.574 & 2.488 \\
WFM-1 db6 & \textbf{0.536} & \textbf{0.703} & \underline{0.786} & 1.441 & 1.949 & 1.025 & 6.732 & 2.927 & 1.553 \\
\midrule
WFM-3 haar & 0.549 & 0.824 & 0.797 & 1.559 & 2.318 & 0.807 & \underline{0.848} & \underline{1.327} & \textbf{1.022} \\
WFM-3 db2 & 0.655 & 0.893 & 0.981 & 1.051 & 1.729 & \textbf{0.766} & 0.933 & 1.349 & 1.085 \\
WFM-3 db4 & 0.586 & 0.759 & 0.803 & \underline{1.029}$^{*}$ & 1.869 & \underline{0.770} & 3.242 & 2.271 & 1.418 \\
WFM-3 db6 & 0.591 & 0.879 & 0.831 & \textbf{0.755} & \underline{1.598} & 0.834 & 2.341 & 2.118 & 1.276 \\
\midrule
FM$_\text{pixel}$ & 0.574 & 0.891 & 0.830 & 1.319 & 1.962 & 0.821 & \textbf{0.580} & 1.380 & 1.028 \\
FM$_\text{fourier}$ & 0.619 & 0.771 & \textbf{0.782} & 1.098 & \textbf{0.810} & 0.839 & 1.302 & \textbf{1.135} & \underline{1.027} \\
WDNO & 1.517 & 1.834 & 1.953 & 19.262 & 8.171 & 1.392 & 4.475 & 2.181 & 1.400 \\
LatentFM$_{\div 64}$ & -- & -- & -- & 9.515 & 4.080 & 0.917 & -- & -- & -- \\
\midrule
WNO & inf & inf & inf & 44.226 & 24.463 & 2.452 & 10.669 & 4.321 & 2.002 \\
FNO & 2.952 & 4.978 & 4.349 & 3.368 & 3.972 & inf & 1.682 & 3.227 & 1.727 \\
TFNO & 2.300 & 6.534 & 6.387 & 102.678 & 31.842 & 2.965 & 4.443 & 2.968 & 1.525 \\
\bottomrule
\end{tabular}
}
\end{table}

\newpage
\paragraph{CRPS across autoregressive rollout steps.} \quad

\begin{table}[ht]
\centering
\caption{CRPS across rollout horizons for all models and PDE systems. LatentFM$_{\div 64}$ is evaluated on Rayleigh--B\'enard only. Mean over three inference seeds (42, 84, 168); the standard deviation across seeds is below 5\% of the mean for all entries except those marked $^{*}$. Here, -- indicates that the model was not available for the corresponding system.}
\label{tab:appendix_crps}
\resizebox{\linewidth}{!}{%
\begin{tabular}{l ccc ccc ccc}
\toprule
 & \multicolumn{3}{c}{Turbulent Radiative Layer 2D} & \multicolumn{3}{c}{Rayleigh--B\'{e}nard} & \multicolumn{3}{c}{Active Matter} \\
\cmidrule(lr){2-4} \cmidrule(lr){5-7} \cmidrule(lr){8-10}
Rollout & 01:32 & 33:64 & 65:97 & 01:32 & 33:64 & 65:196 & 01:16 & 17:32 & 33:77 \\
\midrule
WFM-1 haar & 0.492 & 1.047 & 1.218 & 0.005 & 0.037 & 0.074 & 0.119 & 0.190 & 0.198 \\
WFM-1 db2 & 0.447 & 1.097 & 1.316 & 0.005 & 0.036 & 0.073 & 0.173 & 0.239 & 0.261 \\
WFM-1 db4 & 0.521 & 1.052 & 1.241 & 0.007 & 0.041 & 0.073 & 0.326 & 0.388 & 0.387 \\
WFM-1 db6 & 0.451 & \textbf{0.983} & 1.188 & 0.005 & 0.040 & 0.090 & 0.175 & 0.222 & 0.231 \\
\midrule
WFM-3 haar & \textbf{0.410} & 1.018 & \underline{1.178} & 0.004 & 0.037 & 0.077 & \underline{0.080} & \underline{0.169} & \textbf{0.191} \\
WFM-3 db2 & 0.465 & 1.133 & 1.179 & \textbf{0.003} & 0.032 & \underline{0.069} & 0.087 & 0.176 & 0.196 \\
WFM-3 db4 & 0.427 & \underline{0.984} & \textbf{1.130} & \underline{0.004} & \underline{0.032} & \textbf{0.069} & 0.109 & 0.209 & 0.239 \\
WFM-3 db6 & 0.451 & 1.026 & 1.200 & 0.004 & \textbf{0.031} & 0.079 & 0.100 & 0.197 & 0.213 \\
\midrule
FM$_\text{pixel}$ & \underline{0.418} & 1.088 & 1.210 & 0.005 & 0.036 & 0.076$^{*}$ & \textbf{0.059} & \textbf{0.166} & \underline{0.194} \\
FM$_\text{fourier}$ & 0.857 & 1.418 & 1.560 & 0.009 & 0.038 & 0.096 & 0.185 & 0.227 & 0.218 \\
WDNO & 2.460 & 3.374 & 4.284 & 0.050 & 0.102 & 0.138 & 0.218 & 0.245 & 0.266 \\
LatentFM$_{\div 64}$ & -- & -- & -- & 0.026 & 0.057 & 0.088 & -- & -- & -- \\
\bottomrule
\end{tabular}
}
\end{table}

\newpage
\paragraph{Spectral Coherence RMSE.} \quad

\begin{table}[ht]
\centering
\caption{Spectral coherence RMSE across \textit{Low}, \textit{Mid}, and \textit{High} frequency bands for all models and PDE systems, averaged over rollout horizons. LatentFM$_{\div 64}$ is evaluated on Rayleigh--B\'enard only. Mean over three inference seeds (42, 84, 168); the standard deviation across seeds is below 5\% of the mean for all entries. Here, -- indicates that the model was not available for the corresponding system.}
\label{tab:appendix_spectral}
\resizebox{\linewidth}{!}{%
\begin{tabular}{l ccc ccc ccc}
\toprule
 & \multicolumn{3}{c}{Turbulent Radiative Layer 2D} & \multicolumn{3}{c}{Rayleigh--B\'{e}nard} & \multicolumn{3}{c}{Active Matter} \\
\cmidrule(lr){2-4} \cmidrule(lr){5-7} \cmidrule(lr){8-10}
Frequency Bands & Low & Mid & High & Low & Mid & High & Low & Mid & High \\
\midrule
WFM-1 haar & \underline{0.132} & \underline{0.175} & \textbf{0.134} & 0.171 & 0.132 & 0.116 & 0.229 & 0.291 & 0.204 \\
WFM-1 db2 & 0.139 & 0.181 & 0.142 & 0.167 & 0.126 & 0.105 & 0.291 & 0.336 & 0.236 \\
WFM-1 db4 & \textbf{0.132} & 0.175 & 0.140 & 0.168 & 0.118 & \underline{0.077} & 0.306 & 0.351 & 0.251 \\
WFM-1 db6 & 0.133 & 0.175 & 0.141 & 0.196 & 0.131 & 0.093 & 0.283 & 0.346 & 0.242 \\
\midrule
WFM-3 haar & 0.133 & \textbf{0.174} & 0.141 & 0.149 & 0.130 & 0.114 & 0.205 & 0.280 & 0.256 \\
WFM-3 db2 & 0.139 & 0.188 & \underline{0.136} & \textbf{0.141} & 0.113 & 0.079 & 0.199 & 0.262 & \textbf{0.198} \\
WFM-3 db4 & 0.141 & 0.178 & 0.144 & \underline{0.143} & \textbf{0.109} & \textbf{0.075} & 0.219 & 0.292 & 0.235 \\
WFM-3 db6 & 0.136 & 0.175 & 0.148 & 0.147 & \underline{0.110} & 0.080 & \underline{0.196} & \textbf{0.258} & 0.260 \\
\midrule
FM$_\text{pixel}$ & 0.141 & 0.181 & 0.138 & 0.149 & 0.129 & 0.115 & \textbf{0.183} & \underline{0.258} & 0.291 \\
FM$_\text{fourier}$ & 0.235 & 0.487 & 0.415 & 0.327 & 0.263 & 0.209 & 0.426 & 0.463 & 0.238 \\
WDNO & 0.199 & 0.249 & 0.366 & 0.392 & 0.474 & 0.535 & 0.367 & 0.345 & \underline{0.200} \\
LatentFM$_{\div 64}$ & -- & -- & -- & 0.196 & 0.159 & 0.131 & -- & -- & -- \\
\bottomrule
\end{tabular}
}
\end{table}

\clearpage
\paragraph{Turbulent Radiative Layer 2D: sensitivity to cooling time $t_{\mathrm{cool}}$} \quad

\begin{table}[ht]
\centering
\caption{CRPS per $t_{\mathrm{cool}}$ on TRL, averaged over rollout steps. Mean over three inference seeds (42, 84, 168); the standard deviation across seeds is below 5\% of the mean for all entries except those marked $^{*}$.}
\label{tab:ablation_trl_crps_tcool}
\resizebox{\linewidth}{!}{%
\begin{tabular}{lccccccccc}
\toprule
$t_{\mathrm{cool}}$ & 0.03 & 0.06 & 0.1 & 0.18 & 0.32 & 0.56 & 1 & 1.78 & 3.16 \\
\midrule
WFM-1 haar & 1.165 & 1.135 & 0.865$^{*}$ & 1.044 & 0.830$^{*}$ & 0.798 & 0.806$^{*}$ & 0.983 & 0.673 \\
WFM-1 db2 & 1.112 & 1.078 & 0.878 & 1.057 & 0.880$^{*}$ & 0.779 & 0.853$^{*}$ & 0.976$^{*}$ & 1.002$^{*}$ \\
WFM-1 db4 & 1.170 & 1.102 & 0.847$^{*}$ & 1.084 & 0.875 & 0.849 & 0.879$^{*}$ & 0.946 & 0.719 \\
WFM-1 db6 & 1.076 & 1.074 & \underline{0.831} & 1.041 & 0.823 & 0.773 & 0.741 & \underline{0.943} & \underline{0.594}$^{*}$ \\
\midrule
WFM-3 haar & 1.107 & 1.117 & 0.868 & \underline{0.988}$^{*}$ & \textbf{0.773} & \underline{0.765}$^{*}$ & \textbf{0.675} & 0.961 & 0.595 \\
WFM-3 db2 & 1.248 & 1.141 & 0.894 & 1.027 & 0.836 & 0.783$^{*}$ & 0.751$^{*}$ & 1.019 & 0.656 \\
WFM-3 db4 & \textbf{1.048} & \textbf{1.006} & \textbf{0.817}$^{*}$ & \textbf{0.982} & \underline{0.817} & \textbf{0.749} & \underline{0.692} & \textbf{0.906} & 0.632 \\
WFM-3 db6 & \underline{1.075} & \underline{1.034} & 0.882 & 1.000$^{*}$ & 0.863 & 0.850 & 0.770$^{*}$ & 0.969 & 0.617 \\
\midrule
FM$_\text{pixel}$ & 1.220 & 1.126 & 0.875$^{*}$ & 1.061 & 0.855$^{*}$ & 0.783 & 0.731 & 0.978 & \textbf{0.551}$^{*}$ \\
FM$_\text{fourier}$ & 1.439 & 1.424 & 1.195 & 1.419 & 1.278 & 1.226 & 1.170 & 1.347 & 1.032 \\
WDNO & 3.330 & 3.292 & 3.183 & 3.419 & 3.369 & 3.409 & 3.459 & 3.552 & 3.424 \\
\bottomrule
\end{tabular}
}
\end{table}

\begin{table}[ht]
\centering
\caption{Spectral coherence RMSE per $t_{\mathrm{cool}}$ and frequency band on TRL. Mean over three inference seeds (42, 84, 168); the standard deviation across seeds is below 5\% of the mean for all entries.}
\label{tab:ablation_trl_spectral_tcool}
\resizebox{\linewidth}{!}{%
\begin{tabular}{llccccccccc}
\toprule
Model & $t_{\mathrm{cool}}$ & 0.03 & 0.06 & 0.1 & 0.18 & 0.32 & 0.56 & 1 & 1.78 & 3.16 \\
\midrule
\multirow{3}{*}{WFM-1 haar} & Low & 0.138 & \underline{0.139} & \underline{0.138} & 0.138 & \textbf{0.132} & 0.125 & 0.127 & 0.136 & 0.121 \\
 & Mid & \underline{0.190} & \underline{0.194} & 0.189 & \underline{0.189} & \textbf{0.180} & 0.172 & \underline{0.157} & 0.155 & 0.150 \\
 & High & \underline{0.138} & \underline{0.138} & \textbf{0.136} & \textbf{0.140} & \underline{0.147} & 0.147 & 0.133 & 0.120 & 0.106 \\
\midrule
\multirow{3}{*}{WFM-1 db2} & Low & 0.141 & \textbf{0.139} & 0.140 & 0.137 & 0.143 & 0.134 & 0.136 & 0.134 & 0.149 \\
 & Mid & 0.192 & 0.195 & 0.189 & 0.193 & 0.189 & 0.179 & 0.170 & 0.163 & 0.164 \\
 & High & 0.142 & 0.143 & 0.139 & 0.145 & 0.153 & 0.155 & 0.145 & 0.134 & 0.122 \\
\midrule
\multirow{3}{*}{WFM-1 db4} & Low & \textbf{0.136} & 0.141 & \textbf{0.134} & \textbf{0.134} & 0.136 & 0.130 & 0.127 & \textbf{0.129} & 0.123 \\
 & Mid & \textbf{0.189} & \textbf{0.192} & \textbf{0.185} & \textbf{0.188} & 0.184 & 0.177 & 0.162 & 0.154 & 0.148 \\
 & High & \textbf{0.136} & \textbf{0.138} & \underline{0.137} & 0.144 & 0.151 & 0.158 & 0.147 & 0.135 & 0.112 \\
\midrule
\multirow{3}{*}{WFM-1 db6} & Low & \underline{0.138} & 0.143 & 0.139 & \underline{0.136} & 0.136 & \textbf{0.124} & 0.128 & 0.134 & 0.120 \\
 & Mid & 0.190 & 0.196 & 0.191 & 0.191 & 0.183 & 0.172 & 0.158 & \underline{0.151} & \underline{0.146} \\
 & High & 0.139 & 0.140 & 0.141 & 0.147 & 0.154 & 0.160 & 0.145 & 0.131 & 0.112 \\
\midrule
\multirow{3}{*}{WFM-3 haar} & Low & 0.143 & 0.144 & 0.139 & 0.136 & \underline{0.133} & 0.127 & \underline{0.123} & \underline{0.132} & \textbf{0.118} \\
 & Mid & 0.194 & 0.199 & \underline{0.188} & 0.190 & \underline{0.181} & \textbf{0.167} & \textbf{0.153} & \textbf{0.150} & \textbf{0.144} \\
 & High & 0.152 & 0.152 & 0.148 & 0.148 & 0.157 & 0.152 & 0.135 & 0.123 & \underline{0.103} \\
\midrule
\multirow{3}{*}{WFM-3 db2} & Low & 0.150 & 0.146 & 0.145 & 0.148 & 0.141 & 0.136 & \textbf{0.123} & 0.141 & 0.123 \\
 & Mid & 0.193 & 0.196 & 0.191 & 0.201 & 0.195 & 0.188 & 0.174 & 0.180 & 0.175 \\
 & High & 0.152 & 0.151 & 0.146 & \underline{0.143} & \textbf{0.141} & \textbf{0.140} & \underline{0.127} & \underline{0.120} & 0.105 \\
\midrule
\multirow{3}{*}{WFM-3 db4} & Low & 0.151 & 0.146 & 0.141 & 0.144 & 0.146 & 0.137 & 0.132 & 0.140 & 0.131 \\
 & Mid & 0.195 & 0.197 & 0.189 & 0.192 & 0.189 & 0.176 & 0.162 & 0.155 & 0.147 \\
 & High & 0.149 & 0.152 & 0.150 & 0.154 & 0.159 & 0.157 & 0.141 & 0.129 & 0.109 \\
\midrule
\multirow{3}{*}{WFM-3 db6} & Low & 0.145 & 0.147 & 0.144 & 0.139 & 0.141 & \underline{0.125} & 0.132 & 0.136 & \underline{0.119} \\
 & Mid & 0.193 & 0.198 & 0.188 & 0.191 & 0.181 & \underline{0.168} & 0.158 & 0.152 & 0.147 \\
 & High & 0.148 & 0.150 & 0.151 & 0.156 & 0.170 & 0.164 & 0.144 & 0.134 & 0.112 \\
\midrule
\multirow{3}{*}{FM$_\text{pixel}$} & Low & 0.144 & 0.142 & 0.142 & 0.143 & 0.141 & 0.143 & 0.137 & 0.154 & 0.126 \\
 & Mid & 0.193 & 0.199 & 0.191 & 0.196 & 0.186 & 0.187 & 0.169 & 0.163 & 0.151 \\
 & High & 0.153 & 0.157 & 0.149 & 0.150 & 0.153 & \underline{0.142} & \textbf{0.126} & \textbf{0.113} & \textbf{0.098} \\
\midrule
\multirow{3}{*}{FM$_\text{fourier}$} & Low & 0.242 & 0.245 & 0.228 & 0.249 & 0.238 & 0.229 & 0.229 & 0.243 & 0.217 \\
 & Mid & 0.496 & 0.497 & 0.493 & 0.502 & 0.493 & 0.487 & 0.481 & 0.468 & 0.464 \\
 & High & 0.490 & 0.481 & 0.460 & 0.454 & 0.413 & 0.385 & 0.362 & 0.349 & 0.344 \\
\midrule
\multirow{3}{*}{WDNO} & Low & 0.204 & 0.203 & 0.194 & 0.205 & 0.195 & 0.192 & 0.195 & 0.215 & 0.190 \\
 & Mid & 0.253 & 0.252 & 0.250 & 0.246 & 0.241 & 0.244 & 0.247 & 0.251 & 0.262 \\
 & High & 0.359 & 0.363 & 0.355 & 0.363 & 0.367 & 0.370 & 0.376 & 0.378 & 0.373 \\
\bottomrule
\end{tabular}
}
\end{table}

\clearpage
\paragraph{Rayleigh--Bénard: sensitivity to Rayleigh and Prandtl numbers.} \quad

\begin{table}[ht]
\centering
\caption{CRPS per Rayleigh number $Ra$ on Rayleigh--B\'enard, averaged over Prandtl numbers. Mean over three inference seeds (42, 84, 168); the standard deviation across seeds is below 5\% of the mean for all entries except those marked $^{*}$.}
\label{tab:ablation_rb_crps_ra}
\begin{tabular}{lccc}
\toprule
$Ra$ & $10^{6}$ & $10^{8}$ & $10^{10}$ \\
\midrule
WFM-1 haar & 0.059 & 0.060$^{*}$ & 0.051 \\
WFM-1 db2 & \textbf{0.052} & 0.062$^{*}$ & 0.052$^{*}$ \\
WFM-1 db4 & 0.059$^{*}$ & 0.060 & 0.051 \\
WFM-1 db6 & 0.060$^{*}$ & 0.069$^{*}$ & 0.075 \\
\midrule
WFM-3 haar & 0.060$^{*}$ & 0.060 & 0.055$^{*}$ \\
WFM-3 db2 & 0.054$^{*}$ & \underline{0.055} & \underline{0.049} \\
WFM-3 db4 & \underline{0.053}$^{*}$ & 0.058 & \textbf{0.048}$^{*}$ \\
WFM-3 db6 & 0.065$^{*}$ & \textbf{0.055} & 0.057 \\
\midrule
FM$_\text{pixel}$ & 0.057$^{*}$ & 0.066$^{*}$ & 0.051 \\
FM$_\text{fourier}$ & 0.071 & 0.081 & 0.065 \\
WDNO & 0.120 & 0.117 & 0.116 \\
LatentFM$_{\div 64}$ & 0.087$^{*}$ & 0.067 & 0.065 \\
\bottomrule
\end{tabular}
\end{table}

\begin{table}[ht]
\centering
\caption{CRPS per Prandtl number $Pr$ on Rayleigh--B\'enard, averaged over Rayleigh numbers. Mean over three inference seeds (42, 84, 168); the standard deviation across seeds is below 5\% of the mean for all entries except those marked $^{*}$.}
\label{tab:ablation_rb_crps_pr}
\begin{tabular}{lccc}
\toprule
$Pr$ & $10^{-1}$ & $1$ & $10^{1}$ \\
\midrule
WFM-1 haar & 0.091 & 0.050$^{*}$ & \underline{0.029} \\
WFM-1 db2 & 0.088$^{*}$ & 0.049 & 0.030 \\
WFM-1 db4 & 0.088$^{*}$ & \underline{0.048} & 0.034 \\
WFM-1 db6 & 0.087 & 0.066$^{*}$ & 0.051 \\
\midrule
WFM-3 haar & 0.092$^{*}$ & 0.050 & 0.033 \\
WFM-3 db2 & \underline{0.079} & \textbf{0.048} & 0.031 \\
WFM-3 db4 & \textbf{0.079}$^{*}$ & 0.051$^{*}$ & \textbf{0.028} \\
WFM-3 db6 & 0.086$^{*}$ & 0.050 & 0.040 \\
\midrule
FM$_\text{pixel}$ & 0.088$^{*}$ & 0.054 & 0.033 \\
FM$_\text{fourier}$ & 0.114 & 0.058 & 0.045 \\
WDNO & 0.146 & 0.112 & 0.096 \\
LatentFM$_{\div 64}$ & 0.101$^{*}$ & 0.062 & 0.056 \\
\bottomrule
\end{tabular}
\end{table}

%%%%%%%

% Preamble: \usepackage{subcaption}

\begin{table*}[ht]
\centering
\caption{Spectral coherence RMSE per frequency band on Rayleigh--Bénard, broken down by Rayleigh number $Ra$ (left, averaged over $Pr$) and Prandtl number $Pr$ (right, averaged over $Ra$). Mean over three inference seeds (42, 84, 168); the standard deviation across seeds is below 5\% of the mean for all entries except those marked $^{*}$.}
\label{tab:ablation_rb_spectral}
\begin{subtable}[t]{0.48\linewidth}
    \centering
    \caption{vs $Ra$ (averaged over $Pr$)}
    \label{tab:ablation_rb_spectral_ra}
    \resizebox{\linewidth}{!}{%
    \begin{tabular}{llccc}
    \toprule
    Model & $Ra$ & $10^{6}$ & $10^{8}$ & $10^{10}$ \\
    \midrule
    \multirow{3}{*}{WFM-1 haar} & Low & 0.174 & 0.166 & 0.138 \\
     & Mid & 0.189 & 0.135 & 0.133 \\
     & High & 0.110 & 0.186 & 0.163 \\
    \midrule
    \multirow{3}{*}{WFM-1 db2} & Low & 0.154 & 0.176 & 0.140 \\
     & Mid & 0.161 & 0.132 & 0.134 \\
     & High & 0.092 & 0.163 & 0.150 \\
    \midrule
    \multirow{3}{*}{WFM-1 db4} & Low & 0.160$^{*}$ & 0.169 & 0.140 \\
     & Mid & \underline{0.121} & 0.130 & 0.131 \\
     & High & \textbf{0.049} & 0.102 & \underline{0.120} \\
    \midrule
    \multirow{3}{*}{WFM-1 db6} & Low & 0.199 & 0.219 & 0.210 \\
     & Mid & 0.160 & 0.138 & 0.178 \\
     & High & 0.066 & 0.121$^{*}$ & 0.183 \\
    \midrule
    \multirow{3}{*}{WFM-3 haar} & Low & 0.154 & 0.163 & 0.130 \\
     & Mid & 0.188 & 0.133 & 0.129 \\
     & High & 0.110 & 0.181 & 0.156 \\
    \midrule
    \multirow{3}{*}{WFM-3 db2} & Low & \underline{0.138} & \underline{0.159}$^{*}$ & \textbf{0.127} \\
     & Mid & 0.130 & \underline{0.120} & \textbf{0.125} \\
     & High & 0.060 & 0.104 & \textbf{0.120} \\
    \midrule
    \multirow{3}{*}{WFM-3 db4} & Low & \textbf{0.137}$^{*}$ & 0.160 & \underline{0.130} \\
     & Mid & \textbf{0.118} & \textbf{0.119} & \underline{0.125} \\
     & High & \underline{0.052} & \textbf{0.097} & 0.120 \\
    \midrule
    \multirow{3}{*}{WFM-3 db6} & Low & 0.145 & 0.166 & 0.138 \\
     & Mid & 0.122 & 0.120 & 0.130 \\
     & High & 0.066 & \underline{0.099} & 0.127 \\
    \midrule
    \multirow{3}{*}{FM$_\text{pixel}$} & Low & 0.155 & \textbf{0.150} & 0.131 \\
     & Mid & 0.188 & 0.130 & 0.128 \\
     & High & 0.112 & 0.183 & 0.156 \\
    \midrule
    \multirow{3}{*}{FM$_\text{fourier}$} & Low & 0.347 & 0.321 & 0.299 \\
     & Mid & 0.253 & 0.308 & 0.314 \\
     & High & 0.137 & 0.262 & 0.296 \\
    \midrule
    \multirow{3}{*}{WDNO} & Low & 0.426 & 0.402 & 0.363 \\
     & Mid & 0.549 & 0.484 & 0.448 \\
     & High & 0.585 & 0.581 & 0.529 \\
    \midrule
    \multirow{3}{*}{LatentFM$_{\div 64}$} & Low & 0.206 & 0.183 & 0.167 \\
     & Mid & 0.177 & 0.149 & 0.151 \\
     & High & 0.111 & 0.139 & 0.196 \\
    \bottomrule
    \end{tabular}
    }
    \end{subtable}
\hfill
\begin{subtable}[t]{0.48\linewidth}
    \centering
    \caption{vs $Pr$ (averaged over $Ra$)}
    \label{tab:ablation_rb_spectral_pr}
    \resizebox{\linewidth}{!}{%
    \begin{tabular}{llccc}
    \toprule
    Model & $Pr$ & $10^{-1}$ & $1$ & $10^{1}$ \\
    \midrule
    \multirow{3}{*}{WFM-1 haar} & Low & 0.187 & 0.144 & 0.147 \\
     & Mid & 0.158 & 0.164 & 0.136 \\
     & High & 0.192 & 0.155 & 0.111 \\
    \midrule
    \multirow{3}{*}{WFM-1 db2} & Low & 0.180$^{*}$ & 0.145$^{*}$ & 0.145 \\
     & Mid & 0.150 & 0.153 & 0.123 \\
     & High & 0.182 & 0.136 & 0.087 \\
    \midrule
    \multirow{3}{*}{WFM-1 db4} & Low & 0.181$^{*}$ & 0.141 & 0.148 \\
     & Mid & 0.138 & 0.133 & \underline{0.111} \\
     & High & \underline{0.120} & \textbf{0.087} & \textbf{0.063} \\
    \midrule
    \multirow{3}{*}{WFM-1 db6} & Low & 0.200$^{*}$ & 0.210 & 0.217 \\
     & Mid & 0.148 & 0.162 & 0.165 \\
     & High & 0.129 & 0.150 & 0.091 \\
    \midrule
    \multirow{3}{*}{WFM-3 haar} & Low & 0.171 & 0.139 & 0.137 \\
     & Mid & 0.155 & 0.159 & 0.136 \\
     & High & 0.189 & 0.150 & 0.108 \\
    \midrule
    \multirow{3}{*}{WFM-3 db2} & Low & \textbf{0.160}$^{*}$ & 0.133 & \textbf{0.131} \\
     & Mid & \underline{0.128} & 0.134 & 0.112 \\
     & High & 0.122 & 0.096 & \underline{0.066} \\
    \midrule
    \multirow{3}{*}{WFM-3 db4} & Low & \underline{0.163} & \textbf{0.130} & \underline{0.134} \\
     & Mid & \textbf{0.126} & \textbf{0.128} & \textbf{0.108} \\
     & High & \textbf{0.112} & \underline{0.090} & 0.066 \\
    \midrule
    \multirow{3}{*}{WFM-3 db6} & Low & 0.172 & 0.137 & 0.139 \\
     & Mid & 0.129 & \underline{0.132} & 0.111 \\
     & High & 0.120 & 0.094 & 0.079 \\
    \midrule
    \multirow{3}{*}{FM$_\text{pixel}$} & Low & 0.169 & \underline{0.132} & 0.136 \\
     & Mid & 0.153 & 0.160 & 0.133 \\
     & High & 0.197 & 0.149 & 0.106 \\
    \midrule
    \multirow{3}{*}{FM$_\text{fourier}$} & Low & 0.342 & 0.314 & 0.311 \\
     & Mid & 0.309 & 0.292 & 0.275 \\
     & High & 0.262 & 0.228 & 0.205 \\
    \midrule
    \multirow{3}{*}{WDNO} & Low & 0.390 & 0.388 & 0.413 \\
     & Mid & 0.477 & 0.498 & 0.505 \\
     & High & 0.574 & 0.575 & 0.545 \\
    \midrule
    \multirow{3}{*}{LatentFM$_{\div 64}$} & Low & 0.199$^{*}$ & 0.173 & 0.185 \\
     & Mid & 0.160 & 0.160 & 0.157 \\
     & High & 0.163 & 0.151 & 0.132 \\
    \bottomrule
    \end{tabular}
    }
    \end{subtable}
\end{table*}

\clearpage
\paragraph{Physical consistency: kinetic energy.}\label{si-sec:physics}
To check that the gains in VRMSE and CRPS do not come at the expense of physical realism, we compare the kinetic energy $E_k = \langle \tfrac{1}{2}\rho\,|\mathbf{u}|^2 \rangle$ of each generated sample with that of the reference field at the same step ($\rho = 1$ for RB and AM). The absolute error $|E_k(v_m) - E_k(u)|$ is averaged over the samples and divided by the time-mean reference energy of the trajectory rather than by the instantaneous one, which approaches zero during the early steps of RB. Table~\ref{tab:ke_mae} reports this error per rollout horizon. On TRL, WFM 3-scale matches FM$_\text{pixel}$ up to step 64 and is more accurate at the longest horizon (0.147 vs.\ 0.177). On RB and AM, FM$_\text{pixel}$ has the lowest late-horizon error (0.251 and 0.320, vs.\ 0.348 and 0.531 for WFM 3-scale), while WFM 3-scale remains below FM$_\text{fourier}$ and WDNO on both systems, and below LatentFM$_{\div 64}$ on RB.

\begin{table}[ht]
\centering
\caption{Kinetic-energy relative MAE across rollout horizons (error normalised by the trajectory time-mean of the true kinetic energy). Mean over three inference seeds (42, 84, 168); the standard deviation across seeds is below 5\% of the mean for all entries except those marked $^{*}$. WFM rows use the selected wavelets (Table~\ref{si-tab:optimal_wavelets}). Here, -- indicates that the model was not available for the corresponding system.}
\label{tab:ke_mae}
\resizebox{\linewidth}{!}{%
\begin{tabular}{l ccc ccc ccc}
\toprule
 & \multicolumn{3}{c}{Turbulent Radiative Layer 2D} & \multicolumn{3}{c}{Rayleigh--B\'{e}nard} & \multicolumn{3}{c}{Active Matter} \\
\cmidrule(lr){2-4} \cmidrule(lr){5-7} \cmidrule(lr){8-10}
Rollout & 01:32 & 33:64 & 65:97 & 01:32 & 33:64 & 65:196 & 01:16 & 17:32 & 33:77 \\
\midrule
WFM 1-scale & 0.047 & 0.113 & 0.221 & 0.007 & 0.410 & 0.379 & 0.502 & 0.538 & \underline{0.508} \\
WFM 3-scale & \underline{0.017} & \underline{0.067} & \textbf{0.147} & \textbf{0.003} & \underline{0.217} & \underline{0.348}$^{*}$ & \underline{0.112} & \underline{0.391} & 0.531 \\
\midrule
FM$_\text{pixel}$ & \textbf{0.016}$^{*}$ & \textbf{0.067}$^{*}$ & 0.177 & \underline{0.004} & 0.235 & \textbf{0.251} & \textbf{0.051} & \textbf{0.288} & \textbf{0.320} \\
FM$_\text{fourier}$ & 0.037$^{*}$ & 0.082 & \underline{0.150}$^{*}$ & 0.006 & \textbf{0.153} & 1.018 & 0.440 & 0.513 & 0.642 \\
WDNO & 0.356 & 0.225 & 0.457 & 1.014 & 2.506 & 2.309 & 0.747 & 0.890 & 1.052 \\
LatentFM$_{\div 64}$ & -- & -- & -- & 0.252 & 0.574 & 0.913 & -- & -- & -- \\
\bottomrule
\end{tabular}
}
\end{table}

\paragraph{Robustness to initial-condition perturbations.}\label{si-sec:ic_pert}
We perturb the initial conditions of all TRL test trajectories and roll out WFM 3-scale (haar) as before. Each of the $\mathcal{M}$ trajectories receives independent Gaussian noise on its whole context window, in physical space, with standard deviation $\eta\,\sigma_c$, where $\sigma_c$ is the standard deviation of channel $c$ over the window and $\eta \in \{10^{-3}, 10^{-2}, 10^{-1}\}$. The flow-matching noise is identical across $\eta$, so differences come from the perturbation alone. CRPS changes by at most $2.1\%$ (horizon 65:97, $\eta = 0.1$; Table~\ref{tab:ic_pert_crps}), and spectral coherence RMSE is unchanged at three decimals (Table~\ref{tab:ic_pert_spectral}).

\begin{table}[ht]
\centering
\caption{TRL, WFM 3-scale (haar): CRPS across rollout horizons under Gaussian perturbations of the initial conditions with standard deviation $\eta\,\sigma_c$. Mean over three inference seeds (42, 84, 168); the standard deviation across seeds is below 5\% of the mean for all entries.}
\label{tab:ic_pert_crps}
\begin{tabular}{lccc}
\toprule
IC noise & 01:32 & 33:64 & 65:97 \\
\midrule
$\eta=0$ & 0.410 & 1.018 & 1.178 \\
$\eta=0.001$ & 0.410 & 1.036 & 1.180 \\
$\eta=0.01$ & 0.410 & 1.030 & 1.190 \\
$\eta=0.1$ & 0.413 & 1.026 & 1.203 \\
\bottomrule
\end{tabular}
\end{table}

\begin{table}[ht]
\centering
\caption{TRL, WFM 3-scale (haar): Spectral coherence RMSE across bands under Gaussian perturbations of the initial conditions with standard deviation $\eta\,\sigma_c$. Mean over three inference seeds (42, 84, 168); the standard deviation across seeds is below 5\% of the mean for all entries.}
\label{tab:ic_pert_spectral}
\begin{tabular}{lccc}
\toprule
IC noise & Low & Mid & High \\
\midrule
$\eta=0$ & 0.133 & 0.174 & 0.141 \\
$\eta=0.001$ & 0.133 & 0.174 & 0.141 \\
$\eta=0.01$ & 0.133 & 0.174 & 0.141 \\
$\eta=0.1$ & 0.133 & 0.174 & 0.141 \\
\bottomrule
\end{tabular}
\end{table}

\clearpage
\paragraph{Qualitative Assessment for Active Matter.} \quad

\begin{figure}[h]
  \centering
  %\fbox{\rule[-.5cm]{0cm}{4cm} \rule[-.5cm]{4cm}{0cm}}
  \includegraphics[width=1.0\textwidth]{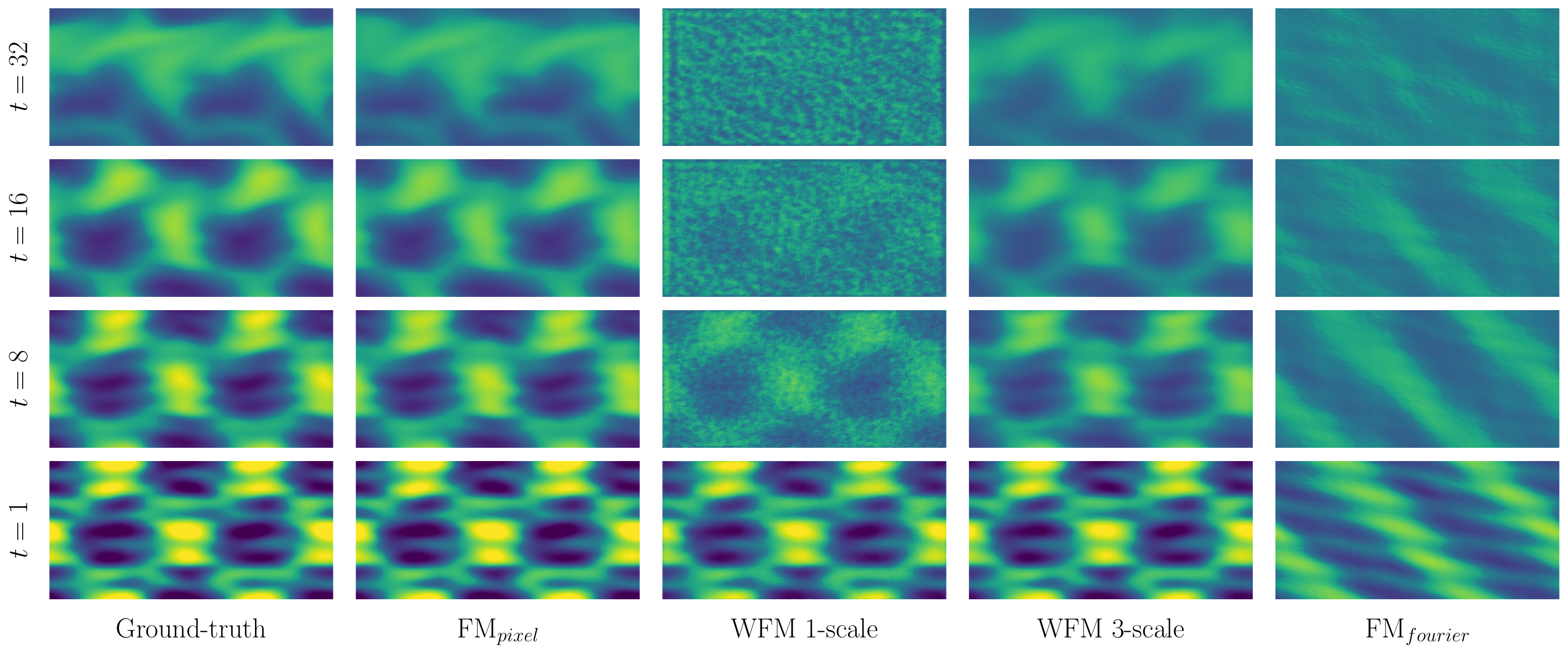}
  \caption{Qualitative rollout comparison for Active Matter: ground truth vs.\ model predictions across autoregressive steps.}
  \label{si-fig:qual_am}
\end{figure}

\clearpage
\section{Profiling Analysis}\label{si-sec:profiling}

\paragraph{Profiling metrics.} Inference wall-clock times are measured by recording system timestamps immediately before the first rollout step and immediately after the last, on the same hardware without concurrent workloads. WFM 1-scale, WFM 3-scale, and FM$_\text{pixel}$ are profiled under identical conditions: $\mathcal{M}=8$ trajectories rolled out for the full test horizon, with each autoregressive step integrating the learned ODE with the Euler method for $N=50$ neural function evaluations (NFEs), for a total of $\mathcal{M} \times T \times N$ NFEs per run. All runs use 4 NVIDIA A100 GPUs on the NSF NCAR Derecho system (\href{https://doi.org/10.5065/qx9a-pg09}{doi:10.5065/qx9a-pg09}).

Timing is collected in a distributed setting with $\mathcal{R}=4$ parallel processes (ranks), one per GPU, each responsible for generating $\lfloor \mathcal{M} / \mathcal{R} \rfloor$ trajectories. To reflect true end-to-end latency, the reported wall-clock time $T_{\text{wall}}$ is taken as the maximum across all ranks:
\begin{equation}
    T_{\text{wall}} = \max_{r \in \{0,\ldots,\mathcal{R}-1\}} T_r,
\end{equation}
where $T_r$ is the local elapsed time on rank $r$, measured with a synchronization barrier both before and after the inference loop to ensure all ranks begin and end counting simultaneously. DataFrame construction, file I/O, and shard merging are explicitly excluded from this measurement.

From $T_{\text{wall}}$, two throughput metrics are derived. The \textit{step throughput} ($\Gamma_{steps}$) measures the total number of rollout steps generated across all ranks per unit time:
\begin{equation}
    \Gamma_{\text{steps}} = \frac{\sum_{r} n_{\text{steps},r}}{T_{\text{wall}}} \quad [\text{steps/s}],
\end{equation}
where $n_{\text{steps},r}$ is the number of rollout steps generated on rank $r$, summed over its trajectories. The \textit{frame throughput} $(\Gamma_{\text{frames}})$ further accounts for spatial resolution by weighting each step by the number of spatial elements $H \times W$:
\begin{equation}
    \Gamma_{\text{frames}} = \frac{\left(\sum_{r} n_{\text{steps},r}\right) \times H \times W}{T_{\text{wall}}} \quad [\text{frames/s}].
\end{equation}
Peak GPU memory consumption (in GiB) is recorded per rank using \texttt{torch.cuda.max\_memory\_allocated} and reported individually to diagnose load imbalance across devices. The \textit{speedup} of a WFM variant relative to $\text{FM}_{\text{pixel}}$ is defined as the ratio of their respective wall-clock times under the same experimental conditions:
\begin{equation}
    \text{Speedup} = \frac{T_{\text{wall}}^{\text{FM}_{\text{pixel}}}}{T_{\text{wall}}^{\text{WFM}}}.
\end{equation}
Since all hyperparameters (number of trajectories, NFEs, ODE solver, hardware) are held fixed across models, any observed speedup is attributable to the multi-scale wavelet space in which the generative process operates. The speedup is moderate and is a by-product rather than the main goal: FM$_{fourier}$, which generates only a truncated set of Fourier modes, has a substantially higher error (Table~\ref{tab:avg_sc_rmse_gen_wavelets}), and latent-space models require pre-training an autoencoder (one day on 8$\times$H100 GPUs for the Rayleigh--B\'enard autoencoder of~\citet{rozet2025}, compared with about six hours on 4$\times$A100 GPUs to train WFM). WFM therefore offers accuracy comparable to or better than pixel-space generation at a lower cost.

\newcolumntype{P}[1]{>{\centering\arraybackslash}p{#1}}

\begin{table}[ht]
\caption{Inference performance across PDEs and models. The reported WFM variants are selected based on the procedure in Section~\ref{sec:wfm_eval}, and summarized in Table~\ref{si-tab:optimal_wavelets}.}
\label{tab:inference_perf_ablation_selected}
\resizebox{\textwidth}{!}{%
\begin{tabular}{ll rrrr rr}
\toprule
\textbf{PDE} & \textbf{Model} & \textbf{Wall Clock (s)} & \textbf{sec / step} & \textbf{GPU Mem (GiB)} & \textbf{Throughput (Mfr/s)} & \textbf{Speedup $\times$} & \textbf{Mem ratio $\times$} \\
\midrule
\multirow{3}{*}{TRL}
  & $\text{FM}_{\text{pixel}}$ & 247.8 & 0.0355 & 0.59 & 1.39 & \textemdash & \textemdash \\
  & WFM 1-scale {\scriptsize($\mathrm{haar}$)} & 198.1 & 0.0284 & 0.59 & 1.73 & 1.25$\times$ & 1.00$\times$ \\
  & WFM 3-scale {\scriptsize($\mathrm{haar}$)} & 218.3 & 0.0313 & 0.66 & 1.57 & 1.14$\times$ & 0.89$\times$ \\
\midrule
\multirow{3}{*}{RB}
  & $\text{FM}_{\text{pixel}}$ & 633.9 & 0.0449 & 1.44 & 1.46 & \textemdash & \textemdash \\
  & WFM 1-scale {\scriptsize($\mathrm{db}4$)} & 418.1 & 0.0296 & 1.44 & 2.21 & 1.52$\times$ & 1.00$\times$ \\
  & WFM 3-scale {\scriptsize($\mathrm{db}4$)} & 457.9 & 0.0324 & 1.51 & 2.02 & 1.38$\times$ & 0.95$\times$ \\
\midrule
\multirow{3}{*}{AM}
  & $\text{FM}_{\text{pixel}}$ & 310.8 & 0.0459 & 1.63 & 1.43 & \textemdash & \textemdash \\
  & WFM 1-scale {\scriptsize($\mathrm{haar}$)} & 213.3 & 0.0315 & 1.64 & 2.08 & 1.46$\times$ & 1.00$\times$ \\
  & WFM 3-scale {\scriptsize($\mathrm{db}2$)} & 247.4 & 0.0365 & 1.71 & 1.80 & 1.26$\times$ & 0.95$\times$ \\
\bottomrule
\end{tabular}%
}
\end{table}

\clearpage
\paragraph{Profiling different WFM representations.}
In this section we ablate the profiling results of WFM variants (1- and 3-scale) across different representations induced by a different choice of mother wavelet. 

\begin{table}[ht]
\caption{Complete inference profiling across all WFM variants (scale and choice of the mother wavelet) relative to FM$_{pixel}$.}
\label{tab:inference_perf_ablation}
\resizebox{\textwidth}{!}{%
\begin{tabular}{ll c cccc cc}
\toprule
\textbf{PDE} & \textbf{Model} & \textbf{Wavelet} & \textbf{Wall Clock (s)} & \textbf{sec / step} & \textbf{GPU Mem (GiB)} & \textbf{Throughput (Mfr/s)} & \textbf{Speedup $\times$} & \textbf{Mem ratio $\times$} \\
\midrule
\multirow{9}{*}{TRL}
  & FM$_{pixel}$ & --- & 247.8 & 0.0355 & 0.59 & 1.39 & \textemdash & \textemdash \\
  \cmidrule(l){2-9}
  & \multirow{4}{*}{WFM 1-scale} & \textbf{haar} & 198.1 & 0.0284 & 0.59 & 1.73 & 1.25$\times$ & 1.00$\times$ \\
  &  & db2 & 199.0 & 0.0285 & 0.59 & 1.72 & 1.24$\times$ & 1.00$\times$ \\
  &  & db4 & 208.1 & 0.0298 & 0.59 & 1.65 & 1.19$\times$ & 1.00$\times$ \\
  &  & db6 & 204.5 & 0.0293 & 0.59 & 1.68 & 1.21$\times$ & 1.00$\times$ \\
  \cmidrule(l){2-9}
  & \multirow{4}{*}{WFM 3-scale} & \textbf{haar} & 218.3 & 0.0313 & 0.66 & 1.57 & 1.14$\times$ & 0.89$\times$ \\
  &  & db2 & 224.0 & 0.0321 & 0.66 & 1.53 & 1.11$\times$ & 0.89$\times$ \\
  &  & db4 & 228.2 & 0.0327 & 0.66 & 1.50 & 1.09$\times$ & 0.89$\times$ \\
  &  & db6 & 221.1 & 0.0317 & 0.66 & 1.55 & 1.12$\times$ & 0.89$\times$ \\
\midrule
\multirow{9}{*}{RB}
  & FM$_{pixel}$ & --- & 633.9 & 0.0449 & 1.44 & 1.46 & \textemdash & \textemdash \\
  \cmidrule(l){2-9}
  & \multirow{4}{*}{WFM 1-scale} & haar & 413.3 & 0.0293 & 1.44 & 2.24 & 1.53$\times$ & 1.00$\times$ \\
  &  & db2 & 417.4 & 0.0296 & 1.44 & 2.22 & 1.52$\times$ & 1.00$\times$ \\
  &  & \textbf{db4} & 418.1 & 0.0296 & 1.44 & 2.21 & 1.52$\times$ & 1.00$\times$ \\
  &  & db6 & 417.4 & 0.0296 & 1.44 & 2.22 & 1.52$\times$ & 1.00$\times$ \\
  \cmidrule(l){2-9}
  & \multirow{4}{*}{WFM 3-scale} & haar & 453.0 & 0.0321 & 1.51 & 2.04 & 1.40$\times$ & 0.95$\times$ \\
  &  & db2 & 446.5 & 0.0316 & 1.51 & 2.07 & 1.42$\times$ & 0.95$\times$ \\
  &  & \textbf{db4} & 457.9 & 0.0324 & 1.51 & 2.02 & 1.38$\times$ & 0.95$\times$ \\
  &  & db6 & 468.0 & 0.0332 & 1.51 & 1.98 & 1.35$\times$ & 0.95$\times$ \\
\midrule
\multirow{9}{*}{AM}
  & FM$_{pixel}$ & --- & 310.8 & 0.0459 & 1.63 & 1.43 & \textemdash & \textemdash \\
  \cmidrule(l){2-9}
  & \multirow{4}{*}{WFM 1-scale} & \textbf{haar} & 213.3 & 0.0315 & 1.64 & 2.08 & 1.46$\times$ & 1.00$\times$ \\
  &  & db2 & 216.9 & 0.0320 & 1.64 & 2.05 & 1.43$\times$ & 1.00$\times$ \\
  &  & db4 & 222.6 & 0.0329 & 1.64 & 1.99 & 1.40$\times$ & 1.00$\times$ \\
  &  & db6 & 228.2 & 0.0337 & 1.64 & 1.95 & 1.36$\times$ & 1.00$\times$ \\
  \cmidrule(l){2-9}
  & \multirow{4}{*}{WFM 3-scale} & haar & 244.7 & 0.0361 & 1.71 & 1.82 & 1.27$\times$ & 0.95$\times$ \\
  &  & \textbf{db2} & 247.4 & 0.0365 & 1.71 & 1.80 & 1.26$\times$ & 0.95$\times$ \\
  &  & db4 & 252.6 & 0.0373 & 1.71 & 1.76 & 1.23$\times$ & 0.95$\times$ \\
  &  & db6 & 261.2 & 0.0386 & 1.71 & 1.70 & 1.19$\times$ & 0.95$\times$ \\
\bottomrule
\end{tabular}%
}
\end{table}

%\newpage
%\clearpage

%\input{checklist.tex}

\end{document}